%% file: main.tex
\pdfoutput=1

\documentclass[11pt]{article}

\usepackage{EACL2023}

\usepackage{times}
\usepackage{latexsym}

\usepackage[T1]{fontenc}

\usepackage[utf8]{inputenc}

\usepackage{microtype}

\usepackage{inconsolata}

\usepackage{enumerate}
\usepackage[colorinlistoftodos]{todonotes}
\usepackage{booktabs}
\usepackage{multirow}
\usepackage{hyperref}
\usepackage{caption}
\usepackage{subcaption}
\usepackage{amssymb}

\title{Instruction Clarification Requests in Multimodal Collaborative Dialogue Games: Tasks, and an Analysis of the CoDraw Dataset}

\author{Brielen Madureira$^\mathbf{1}$ \and David Schlangen$^\mathbf{1,2}$ \\
  $^\mathbf{1}$Computational Linguistics, Department of Linguistics \\
  University of Potsdam, Germany \\
  $^\mathbf{2}$German Research Center for Artificial Intelligence (DFKI), Berlin, Germany \\
  \texttt{\{madureiralasota, david.schlangen}@uni-potsdam.de\} \\}

\begin{document}
\maketitle
\begin{abstract}
In visual instruction-following dialogue games, players can engage in repair mechanisms in face of an ambiguous or underspecified instruction that cannot be fully mapped to actions in the world. In this work, we annotate Instruction Clarification Requests (iCRs) in CoDraw, an existing dataset of interactions in a multimodal collaborative dialogue game. We show that it contains lexically and semantically diverse iCRs being produced self-motivatedly by players deciding to clarify in order to solve the task successfully. With 8.8k iCRs found in 9.9k dialogues, CoDraw-iCR (v1) is a large spontaneous iCR corpus, making it a valuable resource for data-driven research on clarification in dialogue. We then formalise and provide baseline models for two tasks: Determining when to make an iCR and how to recognise them, in order to investigate to what extent these tasks are learnable from data.
\end{abstract}

\section{Introduction} 
    \label{sec:intro}
    \input{contents/intro}

\section{Related Literature}
    \label{sec:lit-review}
    \input{contents/lit-review}

\section{Motivation and Problem Statement}
    \label{sec:problem}
    \input{contents/problem}

\section{Data and Annotation}
    \label{sec:data}
    \input{contents/data}

\section{Corpus Analysis}
    \label{sec:corpus}
    \input{contents/corpus-analysis}

\section{Models and Experiments}
    \label{sec:experiemnts}
    \input{contents/experiments}

\section{Results}
\label{sec:results}
    \input{contents/results.tex}

\section{Discussion}
    \label{sec:discussion}
    \input{contents/discussion}

\section{Conclusion}
    \label{sec:conclusion}
    \input{contents/conclusion}

\section{Limitations}
    \label{sec:limitations}
    \input{contents/limitations.tex}

    
\section*{Acknowledgements}
    We are thankful for the anonymous reviewers for their feedback. We thank our student assistants, Sebastiano Gigliobianco and Sophia Rauh, for performing the annotation and Philipp Sadler for generating the step-by-step scenes.

\bibliography{anthology,custom}
\bibliographystyle{acl_natbib}

\vfill\null

\appendix

\section{Data Statement}
    \label{sec:appendix-data-statement}
    \input{contents/appendix-data-statement}

\section{Additional Corpus Analysis}
    \label{sec:appendix-annotation}
    \input{contents/appendix-annotation}

\section{Reproducibility}
    \label{sec:appendix-reproduc}
    \input{contents/appendix-reproducibility}

\section{Detailed Results}
    \label{sec:appendix-results}
    \input{contents/appendix-results}

\end{document}

%% file: contents/intro.tex
Somewhere in interstellar space are the Voyager Golden Records\footnote{\href{https://voyager.jpl.nasa.gov/golden-record/}{https://voyager.jpl.nasa.gov/golden-record/}⁄}, which left Earth in spacecrafts in 1977 carrying a message about humanity to extraterrestrial civilizations. The committee in charge of designing the message, chaired by Carl Sagan, was careful to include symbolic instructions on how to play the records. But what if these instructions turn out to be incomprehensible to the aliens? 

\begin{figure}
  \includegraphics[trim={0 16cm 19cm 0},clip,width=\linewidth]{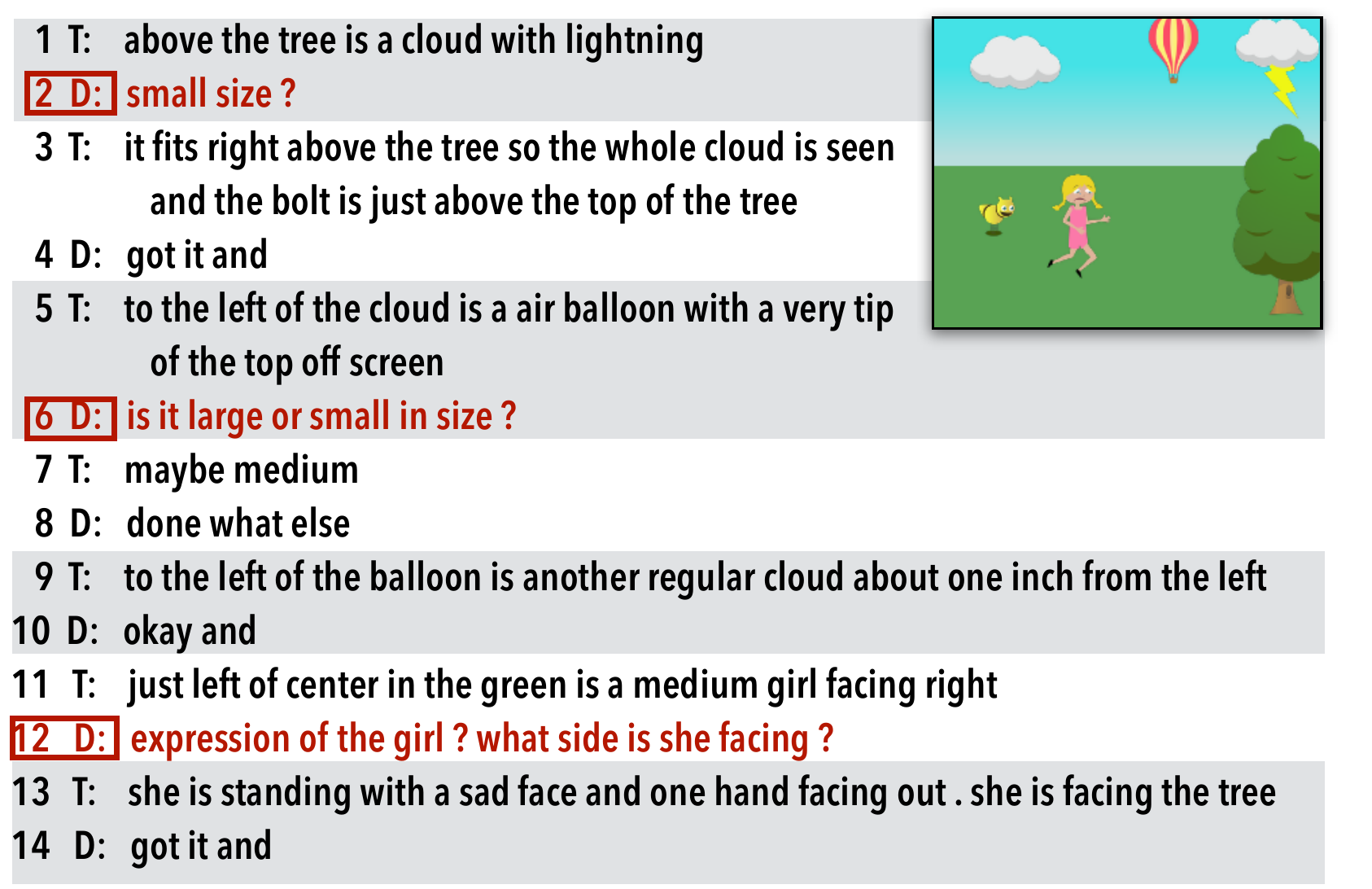}
  \caption{Instruction Clarification Requests identified in a portion of a CoDraw dialogue (ID 8906, \href{https://creativecommons.org/licenses/by-nc/4.0/}{CC BY-NC 4.0}), with a scene from \citet{zitnick2013bringing}.}
  \label{fig:codraw-cr-example}
\end{figure}

In human dialogue, Clarification Requests (CRs), such as those highlighted in Figure~\ref{fig:codraw-cr-example},
are a common and indispensable mechanism to signal misunderstandings and to negotiate meaning, as recently stressed \textit{e.g.}\ by \citet{benotti2017modeling}. This utterance-anaphoric conversational move can be realized with various forms, functions/readings and contents \citep{purver2003means,ginzburg_grounding_2012} and can trigger responses that may or not be satisfactory \citep{rodriguez2004form}.

In addition to the scientific motivation to comprehend CRs as a linguistic phenomenon, timely producing and understanding the vast range of CRs is also a desirable property for dialogue systems \citep{schlangen-2004-causes}. This ability is especially relevant in scenarios where building common ground is necessary to act and collaboratively achieve a goal. Instructional interactions are a particular instance where an instruction follower (\textit{IF}) often needs to ask for clarification in order to execute actions according to an instruction giver's (\textit{IG}) instructions.

Instruction Clarification Requests (iCRs), as we will refer to them, are a type of CRs originating at \citet{clark1996using}'s 4th level of communication, the level of uptake \citep{schloder-fernandez-2014-clarification}. They are elicited when an instruction utterance is generally understood (\textit{e.g.}\ acoustically, syntactically, semantically) but some underspecification or ambiguity prevents the \textit{IF} to carry out an action with enough certainty, as shown in Figure \ref{fig:codraw-cr-example}. 
    
Learning clarification mechanisms from data is still an understudied research problem \citep{benotti-blackburn-2021-recipe}. We envision the following desiderata for a dataset suitable for data-driven research on iCRs:

\begin{itemize}
    \item[$\vartriangleright$] \textbf{Naturalness}: iCRs should occur by the spontaneous decision process of the \textit{IF} in real interaction while trying to act and solve a task, ideally not being induced by external incentives in the data collection and also not synthetically generated.
    \item[$\vartriangleright$] \textbf{Specificity}: the annotation should pin down iCRs as a single category, not subsumed within other CRs and dialogue acts.
    \item[$\vartriangleright$] \textbf{Frequency}: relative and absolute occurrence of iCRs should be large enough for data-driven methods and statistical purposes. 
    \item[$\vartriangleright$] \textbf{Diversity}: iCRs should occur with various forms and content, being grounded in the game actions and parameters.
    \item[$\vartriangleright$] \textbf{Relevance}: iCRs should be pertinent for players to decide on actions and solve the task successfully.
    \item[$\vartriangleright$] \textbf{Regularity}: iCRs should emerge from underlying strategies of the players and not be the result of random or idiosyncratic behaviour.
\end{itemize}

Our research questions are: i) Can \textit{IF} dialogue models trained on data learn to recognise when they would profit from receiving more information in order to execute an action, and thus generate an iCR? ii) Can \textit{IG} dialogue models trained on data learn to recognise when the \textit{IF} is making an iCR and respond to it? 

In this work, our contribution to begin addressing these questions is threefold. We (a) perform annotation of naturally occurring iCRs in a collaborative and multimodal dialogue game, namely the CoDraw dataset \citep{kim-etal-2019-codraw}, showing that it is a valuable resource for data-driven research on clarification in dialogue; (b) analyse the corpus and provide insights relating iCRs to the game dynamics; and (c) discuss two subtasks and models that can be explored with CoDraw-iCR (v1) and may serve as components of \textit{IF} and \textit{IG} dialogue models capable of handling iCRs.

%% file: contents/lit-review.tex
It is a common practice to map CRs to the level of communication \citep{clark1996using,allwood2000activity} where the misunderstanding occurs \citep{gabsdil2003clarification,schlangen-2004-causes,rodriguez2004form,rieser-moore-2005-implications,rieser-etal-2005-corpus,bohus-rudnicky-2005-sorry,benotti-2009-clarification,koulouri-lauria-2009-exploring,benotti-blackburn-2021-recipe}. When ASR used to be a bottleneck for dialogue processing, several works focused on CRs elicited by problems at levels 2 and 3 -- perception and understanding \citep[\textit{inter alia}]{healey2003experimenting,schlangen-fernandez-2007-beyond,schlangen2007speaking,stoyanchev-etal-2013-modelling,stoyanchev2014towards}. Comparatively less research exists focusing on CRs at level 4, namely intention, uptake or task-level clarifications \citep{benotti-2009-clarification,schloder-fernandez-2014-clarification}. We thus contribute to filling this gap, building upon the existing literature we now turn to discuss in more detail.

\citet{schloder-fernandez-2015-clarifying} perform a corpus-based study splitting level 4 CRs into two types of intention-related conversational problems: recognition and adoption. Instruction-following dialogues, where utterances are intertwined with actions, is one setting where level 4 CRs play a fundamental role in negotiating meaning. \citet{benotti2017modeling} discuss the relation between instruction, CRs and contexts in such settings and how  conversational implicatures are a rich source of CRs. Task-level reformulations, a clarification strategy where the initiator rephrases an utterance with respect to its effects on the task, are typically used to confirm more complex actions in instruction giving dialogues \citep{gabsdil2003clarification} and happen very frequently \cite{benotti-2009-clarification}. Multimodality, \textit{e.g.}\ gestures, also play a role in instruction-following CRs \citep{ginzburg-luecking-2021-requesting}.

\citet{benotti-2009-clarification} proposes using planning to infer and generate the task-level clarification potential of instructions and identify level 4 CRs in one dialogue of a corpus of 15 instruction giving dialogues. \citet{benotti-blackburn-2021-recipe} analyse the same corpus and identify six characteristics that may account for the larger proportion of level 4 CRs found in it: task-oriented dialogues, asymmetry in dialogue participant roles (\textit{IF} and \textit{IG}), immediate world validation by the informational or physical actions, shared view and consequent verification of the actions, long dialogues that enable more shared background, and irreversible actions that require more certainty. 

Other corpus studies exist in small datasets. \citet{rodriguez2004form} find that 22.17\% of the CRs are level 4 CRs in an instruction-following setting. Similarly, \citet{gervits-etal-2021-agents} collect and annotate 22 dialogues with a human-controlled virtual robot that followed high-level or low-level instructions. They propose a very detailed annotation schema for the content of CRs, but there is no clear distinction of level 4 CRs.

A larger dialogue game dataset, the Minecraft Dialogue Corpus \citep{narayan-chen-etal-2019-collaborative} with 509 games, has been annotated with CRs. \citet{lambert2019virtual} annotate the \textit{IF} utterances with eight dialogue acts, one of which, \texttt{clarification questions}, comprises requests for clarification to a given instruction or statement (26.36\% of all utterances). \citet{shi-etal-2022-learning} perform a similar annotation with a category \texttt{instruction-level questions} to request clarification for a previous instruction that was not clear or ambiguous (18.64\%). 

The TEACh dataset \citep{padmakumar2022teach} contains 3k dialogues annotated with dialogue acts \citep{gella-etal-2022-dialog}, of which the 675 \texttt{RequestOtherInfo} spans under the \texttt{Instruction} category relate to iCRs. 

\citet{kiseleva2021iglu} extend the Minecraft Dialogue Corpus with 47 games containing 126 CRs for an interactive agent building challenge, but concentrate on the task of modelling a ``silent \textit{IF}'' that cannot ask questions. The second edition of their challenge, which happened recently \citep{kiseleva2022iglu,iglu2022data}, focuses on when the \textit{IF} should ask for clarification and what it should ask about, similar to \citet{aliannejadi-etal-2021-building}. The dataset for the second challenge is not collected through real, synchronous interaction. Instead, one player builds a structure and generates instructions \textit{a posteriori}, and, in a separate step, another player follows these instructions, deciding whether to make a CR. Similarly, \citet{aliannejadi-etal-2021-building} collects a large dataset of CRs to user requests, augmented synthetically, in a multiple-step process without interaction. Another large-scale dataset with 53k task-relevant questions and answers about an instruction was constructed \citet{gao2022dialfred}. However, the data is created by an annotator that does not have to act, but only watches execution videos, asking a question they think would be helpful and then answering their own question.

Although these strategies facilitate data collection, they abstract away the decision-making and repair processes that emerge when humans collaborate to solve a task jointly, which are present in CoDraw. Our work and the existing literature converge in addressing CRs for ambiguous instructions, but CoDraw-iCR (v1) maintains the interactive aspect of \textit{sequential} rounds and the spontaneous initiative of \textit{IF} to ask. It is large in absolute number of iCRs and dialogues, with short games that have a relatively constrained action space. Moreover, our annotation pins down iCRs among other types of CRs.

A dataset that can be further explored for iCRs is \citet{thomason2020vision}. It instantiates a navigation task where the \textit{IF} gets an ambiguous or underspecified command about where to navigate to, and can ask questions to an oracle during the trajectory.

In HRI, following commands is a central task. \citet{koulouri-lauria-2009-exploring} investigate miscommunication management mechanisms in robots performing collaborative tasks, in which task-level reformulations is a challenging type of CR that requires identification of the effects of all possible executions of an instruction. \citet{deits-2013-clarifying} evaluate various clarification question strategies for robots that receive instructions with an ambiguous phrase. \citet{marge-rudnicky-2015-miscommunication} examine recovery strategies in situated grounding problems, when an agent has to deal with requests containing referential ambiguity or that are impossible to execute. Interestingly, \citet{jackson2018robot} and \citet{jackson2019language} raise awareness to the fact that merely posing a CR can already imply willingness to follow a command, which is undesirable in morally delicate situations.

Other tangent research areas study clarification edits to solve underspecified phrases in instructional texts \citep{roth-etal-2022-semeval} and clarification responses in community forum questions or search queries \citep{braslavski2017cqa,rao-daume-iii-2018-learning,aliannejadi2019asking,kumar-black-2020-clarq,hu-etal-2020-interactive,majumder-etal-2021-ask}, scenarios with only minimal or no interaction.\\

\noindent \textbf{Tasks}. Deciding when to initiate a CR in various contexts is a task classically discussed in the CR literature \citep[\textit{inter alia}]{rieser-lemon-2006-using,stoyanchev2012clarification,stoyanchev-etal-2013-modelling,narayan-chen-etal-2019-collaborative,aliannejadi-etal-2021-building,shi-etal-2022-learning,kiseleva2022iglu}. Fewer works exist specifically about detecting if a CR was made. Identification of CRs in corpora carry out a similar task, although this is not done from the perspective of an agent knowing that it needs to respond to the CR, of which \citet{de-boni-manandhar-2003-analysis} is an example. More generally, this task can be subsumed by dialogue act classification, as in, for instance, \citet{gella-etal-2022-dialog}.

%% file: contents/problem.tex
CRs occur naturally in human-human interaction and thus also in visual dialogue games. Neural network-based dialogue models trained at such datasets need to properly handle this phenomenon, which comprises various component tasks for identifying, interpreting, generating and responding to CRs. In this section, we formalise the setting and two of these tasks.

\subsection{Formalisation: Instruction-Following Dialogue Games}
\label{sec:formalisation}

    A visual instruction-following dialogue game can be formalised as a tuple $G = (P, S, R, M)$ representing a goal-oriented interaction between players $P$ (an instruction giver \textit{IG} and an instruction follower \textit{IF}). \textit{IG} sees a scene $S$, hidden to \textit{IF}, and instructs \textit{IF} on how to reconstruct it. They exchange a sequence $R$ of $n$ rounds $r_i=(g_i, a_i, f_i)$ comprised of two utterances $(g_i, f_i)$, from \textit{IG} and \textit{IF}, respectively, and of actions $a_i$ that incrementally create partial reconstructions $s_i$ of $S$. $R$ is initialised as an empty set and, at each round, it is extended with $g_i$, $a_i$ and $f_i$, in that order. The final state of a completed game contains all filled rounds. A scene similarity metric $M$ computes how close the reconstructions are to the original image at each round, and the goal is to maximize similarity of the final reconstruction $M(S, s_n)$.
	
	The dialogue acts by the \textit{IF} include acknowledgements and clarification requests, whereas the dialogue acts by the \textit{IG} include instructions and responses to clarifications. Two variations are possible: the state $s_i$ can be accessible for the \textit{IG} or not. The incremental scenes can be regarded either as the common ground between players (if both can see it) or as what the \textit{IF} considers to be their common ground (when it is private), akin to what is proposed by \citet{mitsuda-etal-2022-dialogue}.

    Following \citet{clark1996using}, we assume that a pair of equally competent players, committed to the game's goal of maximizing $M(S, s_n)$, seek to minimize joint effort. It is acceptable for the \textit{IG} to produce an underspecified instruction if producing a fully specified instruction would cost more than answering an iCR. Instruction CRs require an extra effort by the \textit{IF}, so they should occur when repair is necessary and the cost of asking is lower than the potential information gain.

\subsection{Tasks}
\label{tasks}
	
	We propose to use CoDraw-iCR (v1) to advance research in iCRs by modelling two CRs subtasks in an instruction-following dialogue game grounded in a visual modality. Both subtasks can be regarded as a binary decision step happening right before each player's next utterance generation.\\
 
	\noindent \textbf{Task 1: Ask iCR?} From the \textit{IF}'s perspective as the CR initiator, decide when to initiate a CR. More specifically, after each \textit{IG} utterance, given the dialogue context $D_{0:(i-1)}$ (that is, all previous utterances), the current utterance $g_i$ by $IG$, and the current state\footnote{Under the assumption that the \textit{IF} has manipulated the scene in response to \textit{IG} already. For CoDraw, the exact point when the \textit{IF} types the message has not been preserved.} of the scene $s_{i}$, the \textit{IF} must decide  on the type of their utterance $f_i$, namely whether to consider the action completed and signal willingness to receive further instructions (\textit{e.g.}, produce something like ``OK''), or to ask for clarification on some aspect of a previous instruction. That is, this formulation of the task focuses on the dialogue act to perform, abstracting away from the concrete realisation. It deals with the problem of automatically determining what is a good instruction and what is not, on its context. This task relates to slot filling in the sense that an instruction containing all the needed parameters for the mentioned objects should not require clarification.\\

	\noindent \textbf{Task 2: Was this an iCR?} From the \textit{IG}'s perspective as the CR recipient, identify whether an iCR has been made. At each round $i$, given the dialogue context $D_{0:i}$ (in which the last utterance, $f_i$, is possibly an iCR) and the original scene $S$, the \textit{IG} must decide whether to give further instructions or to (also) respond to an iCR.

%% file: contents/data.tex
CoDraw \cite{kim-etal-2019-codraw} is a collaborative instruction-following dialogue game, in which a ``teller'' (in our terminology, the \textit{IG}) observes a clipart scene and instructs a ``drawer'' (\textit{IF}), who has no access to it, on how to reconstruct it, \textit{i.e.}\ place cliparts in a canvas with the correct size, direction and position. The corresponding crowdsourced dataset contains 9,993 dialogues in English and has been released under a \href{https://creativecommons.org/licenses/by-nc/4.0/}{CC BY-NC 4.0} license. This dataset instantiates the formalisation proposed in Section \ref{sec:problem}, but adds an additional signal: The teller is allowed to peek at the drawer's canvas once during the game whenever they want, \textit{i.e.}\ the teller can get access to $s_i$ and thus judge how it differs from $S$. Players exchange messages of up to 140 characters through a chat interface and must alternate turns. We will use round to refer to a pair of consecutive utterances by teller and drawer with the corresponding actions. The drawer's performance is evaluated with a scene similarity score that ranges from 0 to 5, where 5 is a perfect match. Table \ref{table:stat-codraw} summarizes quantitative aspects of the dataset.

        \begin{table}[h!]
            \centering
            \small
            \begin{tabular}{p{3.1cm} c c c} 
             \toprule
               & \textbf{train} &  \textbf{val} & \textbf{test}  \\ 
             \midrule
                dialogues & 7,989 & 1,002 & 1,002 \\
                \hspace{0.2cm} with peek  & 7,315 & 923 & 913 \\
                avr. final score & 4.20 & 4.19 & 4.17 \\
                \hspace{0.2cm} before peek & 3.97 & 3.95 & 3.96 \\
                avr. rounds/dialogue & 7.76 & 7.69 & 7.70 \\           
                avr. utterance len teller & 14.36 & 14.48 & 14.31 \\  
                avr. utterance len drawer & 2.58 & 2.67 & 2.58 \\
            \midrule
                vocab size \textit{IG} & \multicolumn{3}{c}{4,506} \\
                vocab size \textit{IF} & \multicolumn{3}{c}{2,200} \\
            \bottomrule
            \end{tabular}
            \caption{Descriptive statistics: CoDraw dataset.}
            \label{table:stat-codraw}
        \end{table}

Each game is about a different abstract scene\footnote{\href{http://optimus.cc.gatech.edu/clipart/}{http://optimus.cc.gatech.edu/clipart/}} composed of between 6 and 17 out of a set of 58 clipart types \citep{zitnick2013bringing,zitnick2013learning}, among which the boy and the girl can have 5 facial expressions and 7 body poses, so the resulting clipart set contains 126 elements and the default background. 
Multiple types of trees, hats, clouds, glasses and balls can introduce the need for ambiguity resolution in the games. As the individual components can be placed freely, the space of possible resulting scene images is practically unlimited in size.

In the baseline models proposed in the original paper, the authors introduce a simplifying assumption which removes the drawer's utterances from the dialogue history (they call this condition the \textit{silent drawer}).
The authors leave the tasks of identifying when a CR is necessary and of generating it for future work. Subsequent works with this dataset have focused on text-to-image generation \citep{ElNouby2019TellDA,Matsumori2021LatteGANVG,Zhang2021TextAN,Lee2021VisualTO,Liu2020IRGANIM,Fu2020IterativeLI} but, to the best of our knowledge, no other work has examined CRs in CoDraw. We thus take up this idea to bring back the dialogue modality to this dialogue game.

\noindent \textbf{Identification of Instruction CRs}. We observe that a good portion of the drawer's utterances belongs to one of two dialogue act types: \textit{acknowledgements}, signaling that the teller may proceed with the next instruction, and \textit{clarification requests}, initiating repair on aspects necessary to solve the task. We thus consider CoDraw to be a potentially interesting source of iCRs. 

The first step we take is identifying instruction-level CRs in this dataset. To achieve that, we perform a binary decision over the drawer's utterances. For our purposes, an utterance is an iCR if the following assertion is likely true: 

\begin{itemize}
    \item[] ``\textit{This utterance indicates that the drawer is requesting further information about one or more instruction(s) previously given by the teller in order to perform an action accordingly, likely because part of the instruction was underspecified, ambiguous or not clear.}''
\end{itemize}

To reduce the annotation workload, we annotate utterance \textit{types}; forms that occur only once (88.97\% of the types) are presented with a one-utterance context window around it. All occurrences of each of the other utterance forms are collapsed into a single datum, presented to the annotators without context.

%% file: contents/corpus-analysis.tex
    \begin{figure*}[ht!]
    	\includegraphics[trim={0cm 0.8cm 0cm 0cm},clip,width=\textwidth]{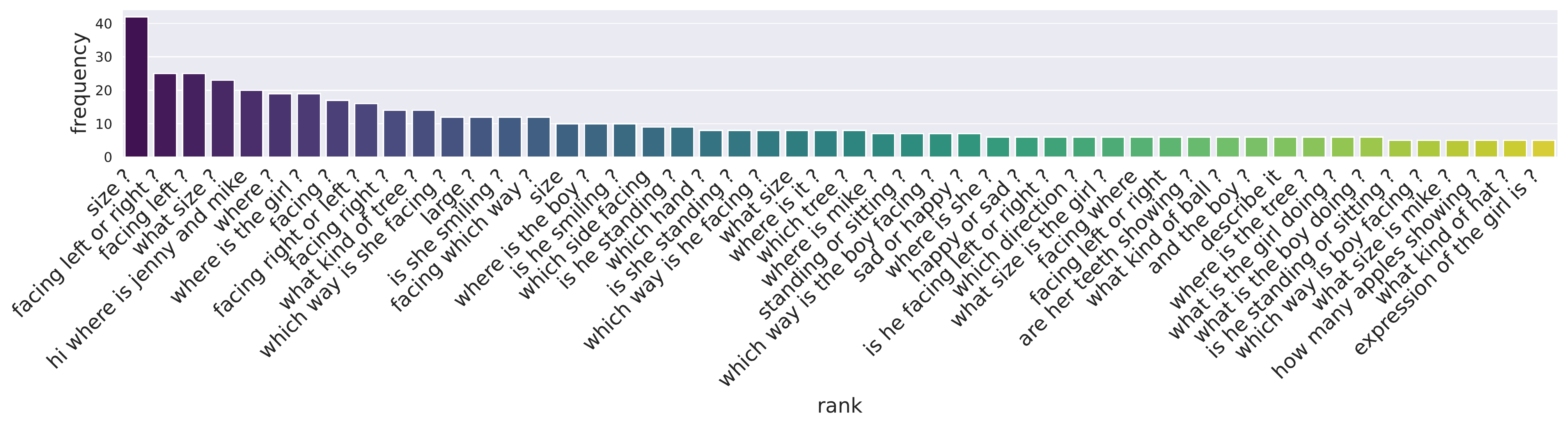}
    	\vspace{-0.8cm}
    	\caption{50 most frequent Instruction CRs in the CoDraw dataset ordered by rank.}
    	\label{fig:common-CRs}
    \end{figure*}

In this section, we present an analysis of iCRs in the CoDraw dataset and their relation to the game dynamics, establishing connections to the items in our desiderata and showing that CoDraw-iCRs (v1) is a promising resource to study the phenomenon and to model dialogue agents that learn what to do in face of unclear instructions, complementing existing initiatives.\footnote{The dataset is available for the community upon request.}

\subsection{Descriptive Statistics}

    \begin{figure*}[ht]
      \includegraphics[trim={0cm 0cm 0cm 0cm},clip,width=\textwidth]{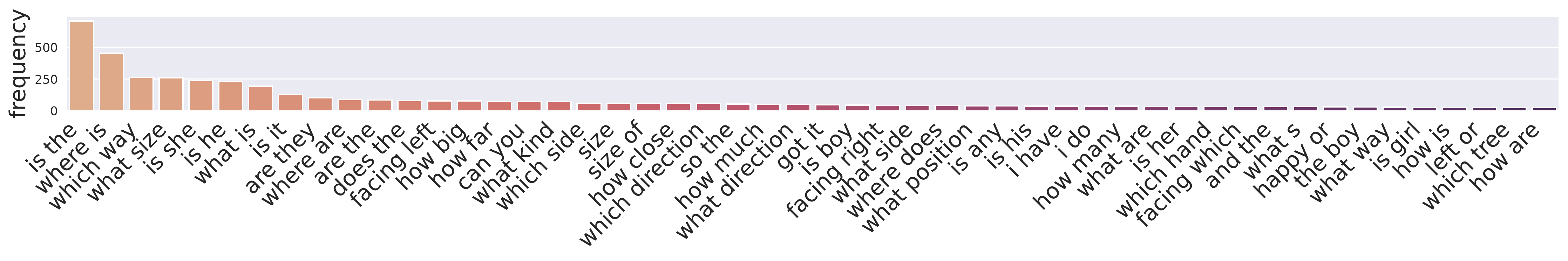}
      \vspace{-0.8cm}
      \caption{50 most frequent iCRs initial bigrams in the CoDraw dataset.}
      \label{fig:initial-bigrams}
    \end{figure*}

    The 13,727 \textit{IF}'s utterance types have been annotated by two annotators, with a Cohen's $\kappa$ \citep{cohen-kappa} of 0.92. Table \ref{table:stat-annotation} presents the main descriptive statistics of the annotated corpus.\footnote{In this paper, for the around 3.6\% of the utterances with disagreements, we opt for the second annotator's labels, who had more training.} 8,807 (11.36\%) of all drawer's utterances in CoDraw are iCRs. 59.45\% of the dialogues contain no iCRs. For the purpose of analysis, we also compute numbers relative to the subset of dialogues that contain at least one iCR; the idea here is that this excludes players who may not have been willing to use the opportunity to ask iCRs. In this subset, the percentage of iCRs is 24.36\%. We also separate out numbers computed from the dialogues up the ``peek'' action described above, as from that move on, the state of the common ground changes.

    \begin{table}[h!]
        \centering
        \small
        \begin{tabular}{r c c c} 
         \toprule
                               & all       &  w/ iCRs  & until peek \\
         \midrule
            dialogues          & 9,993     & 4,052     & - \\
            rounds             & 77,502    & 36,149    & 61,829 \\
            iCR utterances     & 8,807     & 8,807     & 7,803 \\\
            \% iCR utterances  & 11.36     & 24.36     & 12.62 \\
            mean iCRs/dialogue & 0.88      & 2.17      & 0.78 \\
            std iCRs/dialogue  & 1.53      & 1.73      & 1.36 \\        
        \bottomrule
        \end{tabular}
        \caption{Descriptive statistics: Annotation.}
        \label{table:stat-annotation}
    \end{table}

    Figure \ref{fig:common-CRs} presents the most frequent iCR utterance types, ordered by rank. 7,260 (94.13\% of the types) are \textit{hapax legomena}. Types occupying the highest ranks relate to size, position and orientation, which directly map to the possible actions on cliparts, and to disambiguation of \textit{e.g.}\ facial expression and body pose. Few types occur more than 5 times, which is evidence that the dataset contains a rich diversity of iCR surface forms. Figure \ref{fig:initial-bigrams} aggregates iCRs by initial bigrams, after removing punctuation and initial \textit{ok} and \textit{okay} tokens (which realise a different dialogue act). Common iCR forms are polar questions and wh-questions also related to the main actions (placement, resize, flip, disambiguation).
    
\begin{figure}[!ht]
        \begin{subfigure}{\columnwidth}
             \centering
             \includegraphics[trim={0cm 0cm 0cm 1cm},clip,width=\linewidth]{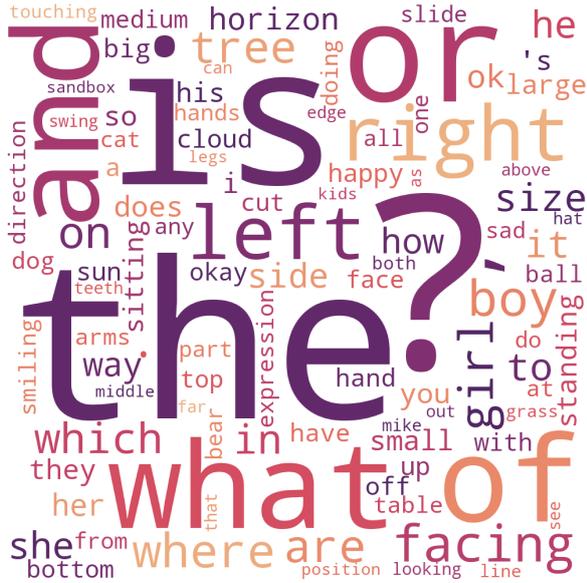} %
             \caption{iCR utterances}
         \end{subfigure}
         \par\bigskip
        \begin{subfigure}{\columnwidth}
             \centering
             \includegraphics[trim={0cm 0cm 0cm 1cm},clip,width=\linewidth]{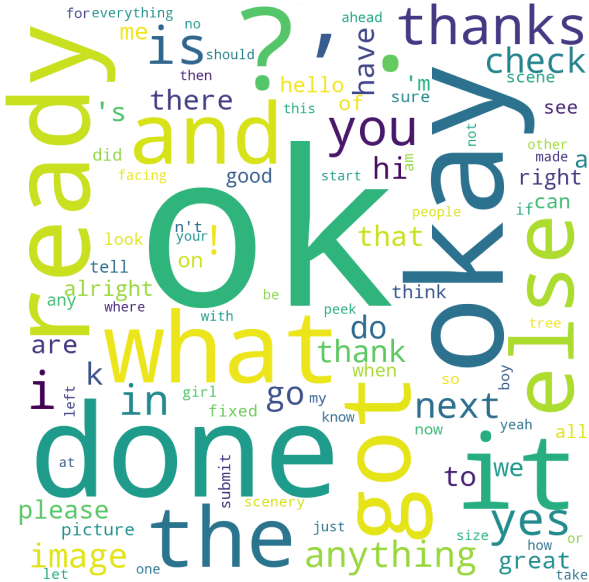}%
             \caption{other utterances}
         \end{subfigure}
        \caption{Most common tokens weighted by frequency.}
        \label{fig:vocab-wordcloud}
    \end{figure}

    The drawer's vocabulary contains 2,200 token types, out of which 1,468 occur in iCRs. Figure \ref{fig:vocab-wordcloud} shows an overview of the 100 most common tokens. The frequent iCR vocabulary contains many nouns relating to cliparts (slide, table, bear, dog), in particular those that refer to nouns involving ambiguity (boy, girl, cloud, tree, ball). Question words occur frequently (what, how, where, which) as well as words about object placement (horizon, facing, size, top, touching, edge). Non-iCR utterances commonly contain words related to the task (scenery, picture, image, check, next), greetings and thanks, and acknowledgement words (ok, ready, done). 

\subsection{Relations to Game Dynamics}

    We now turn to examining how the occurrence of iCRs relate to the overall game dynamics.

    To analyse CRs, three positions in a dialogue are particularly relevant: the source utterance in which the communication problem occurs, the CR utterance where repair is initiated, and the response utterance where the problem should ideally be dealt with. Since the dialogue is organized into a sequence of rounds with pairs of utterances $(g_i, f_i)$, if an iCR occurs at round $i$, then $f_i$ is an iCR, $g_i$ is the likely source utterance, and $g_{i+1}$ is possibly the response utterance. In Figure \ref{fig:codraw-cr-example}, turns 1, 5 and 11 are sources, 2, 6 and 12 are iCRs and 3, 7 and 13 are responses. However, these events do not necessarily occur in immediate sequence.

    Here, we investigate how the game dynamics change at two positions: iCR rounds and rounds immediately following an iCR. We look at the mean number of actions per round and the difference in the score metric with respect to the previous state, as shown in Table \ref{table:stat-game-rounds}. On average, more actions occur at iCR rounds than at non-CR rounds. The difference is even larger in post-iCR rounds, where necessary edits can be occurring. iCR rounds also cause an average higher improvement in the metric than other rounds and the same occurs for rounds after iCRs in dialogues containing iCRs. 
    
    \begin{table}[h!]
        \centering
        \small
        \begin{tabular}{p{3.3cm} l l} 
         \toprule
           &\textbf{all} & \textbf{w/ iCRs} \\ 
         \midrule
         \multicolumn{3}{l}{\textbf{mean actions per round}}  \\

           \hspace{0.1cm} iCR rounds & 1.72 & 1.72 \\
           \hspace{0.1cm} not iCR rounds & 1.64* & 1.62* \\
           \hspace{0.1cm} post-iCR rounds & 2.11 & 2.11 \\
           \hspace{0.1cm} not post-iCR rounds & 1.59* & 1.50* \\
         
         \midrule
         \multicolumn{3}{l}{\textbf{mean score diff} } \\

           \hspace{0.1cm}  iCR rounds & 0.59 & 0.59 \\
           \hspace{0.1cm}  not iCR rounds & 0.53* & 0.43* \\
           \hspace{0.1cm}  post-iCR rounds & 0.53 & 0.53 \\
           \hspace{0.1cm}  not post-iCR rounds & 0.54 & 0.44* \\
         
         \bottomrule
        \end{tabular}
        \caption{Round dynamics. * means the difference in relation to the value at the row above is statistically significant at $\alpha=0.01$ using a permutation test.}
        \label{table:stat-game-rounds}
    \end{table}

    To conclude this section, we refer back to our desiderata. The \textbf{naturalness} of iCRs is a consequence of the data being produced by synchronous human-human interaction in a setting that does not directly induce players to ask for clarification; indeed, almost 60\% of the games do not contain iCRs, which we take to be evidence that they are a result of the private decision making of the \textit{IF} and not due to them following instructions on which dialogue acts to produce. \textbf{Specificity} is guaranteed by the annotation process which had a definition to distinguish iCRs from other utterances. In terms of \textbf{frequency}, iCRs are a common phenomenon in CoDraw-iCR (v1), which contains 8,807 (11.36\%) iCR utterances, a sample larger than existing annotated datasets. We have gathered evidence that \textbf{diversity} is present, given that iCRs occur in various forms and exhibit lexical and semantic variety on content related to the game. When it comes to \textbf{relevance} to the task, we have shown that there are statistically significant differences in number of actions and score differences at turns realising and following iCRs, which is a sign that agents need to process iCRs in order to act accordingly throughout the game. \textbf{Regularity} is addressed in the experiments in the next section.

%% file: contents/experiments.tex
In this section, we present the models for the two tasks discussed in Section \ref{tasks} as well as the evaluation metrics. Both are binary classification tasks using regression to predict the probability of the positive label (iCR) on imbalanced datasets, whose distribution is shown in Table \ref{table:icr-labels}.

        \begin{table}[h!]
            \centering
            \small
            \begin{tabular}{r c c c} 
             \toprule
                 & \textbf{train} & \textbf{val} & \textbf{test} \\ 
             \midrule
              datapoints & 62,067 & 7,714 & 7,721 \\
              \% iCR     & 11.30  & 11.92 & 11.28 \\
              \% not iCR & 88.69  & 88.07 & 88.71 \\
            \bottomrule
            \end{tabular}
            \caption{Distribution of labels.}
            \label{table:icr-labels}
        \end{table}

\subsection{Models}

We model the two prediction subtasks as a function $f: (s, c, u)\mapsto P(l=1)$ where $s$ is the representation of the scene, $c$ is the representation of the dialogue context, $u$ is the representation of the last utterance and $l$ is the label. This function is approximated with a neural network that takes each input embedding, encodes them, and maps them to a concatenated representation which is fed into a two-layer classifier that outputs the probability of the positive label by applying the sigmoid function to the logit output, as illustrated in Figure \ref{fig:models}.\footnote{Details about the implementation, setup and experiments are in the Appendix and the code is available at \href{https://github.com/briemadu/codraw-icr-v1/}{https://github.com/briemadu/codraw-icr-v1/}.}

    \begin{figure}[!ht]
        \begin{subfigure}{\columnwidth}
             \centering
             \includegraphics[trim={0cm 8cm 1cm 0cm},clip,width=\linewidth,page=1]{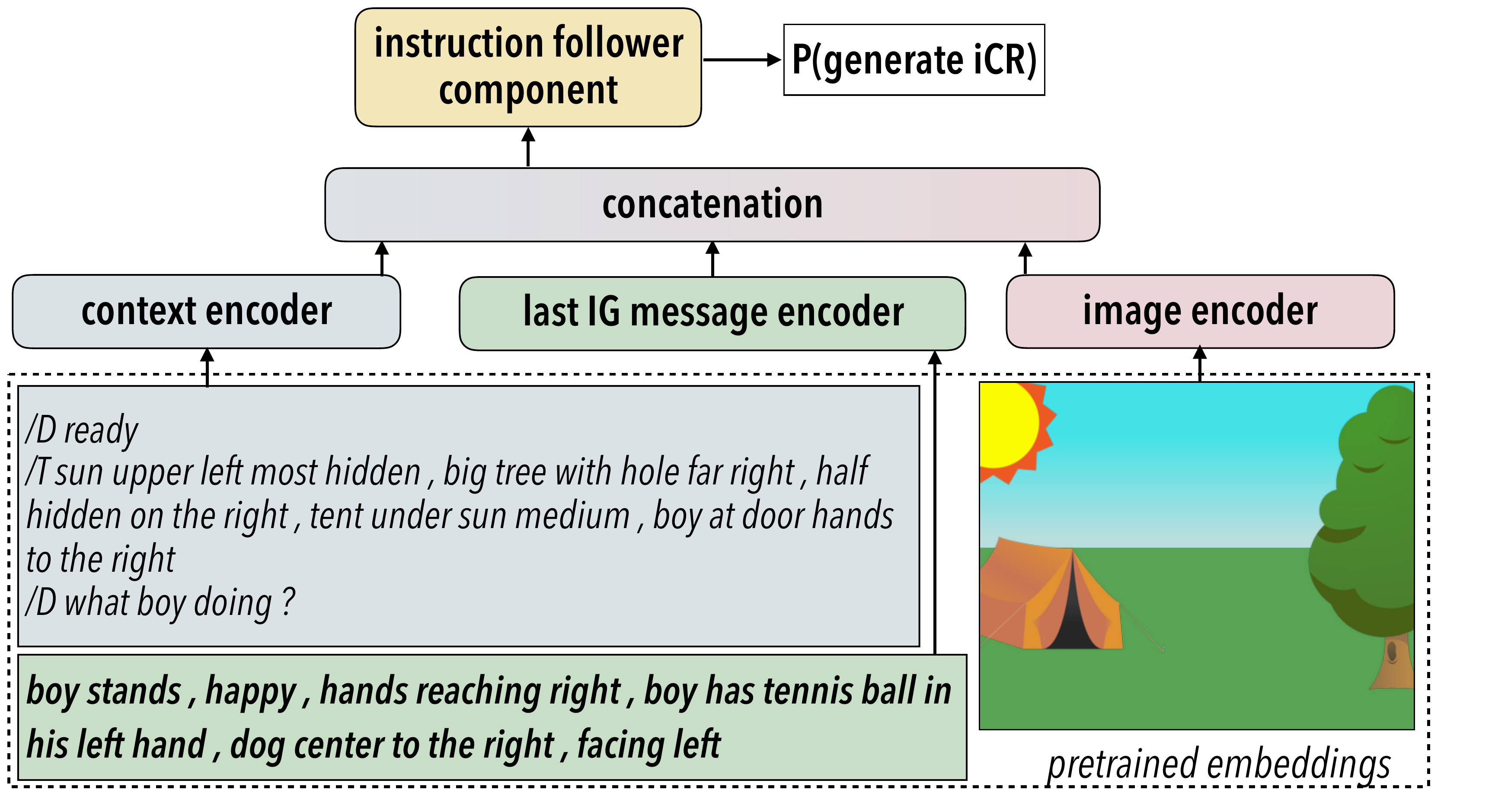} %
             \caption{Task 1: Ask iCR?}
         \end{subfigure} \par\bigskip
        \begin{subfigure}{\columnwidth}
             \centering
              \includegraphics[trim={0cm 8cm 1cm 0cm},clip,width=\linewidth,page=2]{figures/model.pdf}%
             \caption{Task 2: Was this an iCR?}
         \end{subfigure}
        \caption{Illustration of the classifier architecture, with an example dialogue from CoDraw (ID 3454).}
        \label{fig:models}
    \end{figure}

\subsection{Evaluation}

Although the area under the ROC curve is a standard evaluation metric for binary classification, it can be deceptive in imbalanced datasets due to the interpretation of specificity, in which case Precision-Recall curves are more suitable \citep{saito2015precision}. The Average Precision (AP) summarizes this curve into one metric that ranges from 0 to 1, where 1 is the best performance, and the theoretical random is the fraction of positive labels. To facilitate comparison to existing literature, we also report macro-average F1 Score.

As trivial baselines, we perform logistic regression on basic features of the utterances and on the input representation vectors. For Task 1, the features are the length of the last teller's utterance and its boolean bag-of-words representation. For Task 2, we use the length of the last drawer's utterance and a binary variable indicating whether a content word occurs in it. The list of content words was extracted manually from a sample of dialogues.

\subsection{Embeddings}

The pretrained embeddings for texts are generated with SentenceTransformers \citep{reimers-2019-sentence-bert} and for images with ResNet101 \citep{he2016deep}. In order to probe whether the pretrained sentence encoders minimally capture the necessary information for our task, we use the dialogue context representation at the turn before the peek action to predict whether iCRs occurred in the dialogue so far. Using a logistic regression model on dialogues that contain a peek turn, we achieve AP$=0.91$ and macro F1 Score$=0.86$ in the validation set. This provides evidence that, despite they having been optimized for other tasks, the occurrence of iCRs is, to some extent, encoded in the representations.

%% file: contents/results.tex
Table \ref{table:main-results} presents the main results of our models on the two tasks. The feature-based baselines provide some gain over the random performance for Task 1, and a considerable improvement for Task 2. The logistic regression baseline is enough to produce good results for Task 2, whereas Task 1 remains very challenging even for the neural network model.

\begin{table}[h!]
    \centering
    \small
    \begin{tabular}{r r c c c c} 
    \toprule
              &     & \multicolumn{2}{c}{\textbf{Task 1: \textit{IF}}} &  \multicolumn{2}{c}{\textbf{Task 2: \textit{IG}}} \\
    \cmidrule(lr){3-4} \cmidrule(lr){5-6}
       &  & AP    & mF1 & AP    & mF1 \\
     \cmidrule{1-6} 
     \multirow{2}{*}{random}    & val & .117 & .489 & .117 & .489 \\
                                & test & .113 & .503 & .113 & .503 \\ 
     \cmidrule(r){2-6}                
     \multirow{2}{*}{features}  & val & .206 & .531 & .687 & .858 \\ 
                                & test & .195 & .518 & .687 & .855 \\  
      \cmidrule(r){2-6}    
      \multirow{2}{*}{log-reg}  & val & .324 & .587 & .984 & .962 \\ 
                                & test & .287 & .576 & .978 & .961 \\ 
    \cmidrule(r){2-6}    
    \multirow{2}{*}{model}      & val & .399 & .662 & .991 & .969 \\ 
                                & test &  \textbf{.347} & \textbf{.645} & \textbf{.988} & \textbf{.968} \\
    \bottomrule
    \end{tabular}
    \caption{Main results. Average Precision and macro-average F1 Score on the validation and test sets.}
    \label{table:main-results}
\end{table}

\noindent \textbf{Ablation.} We remove each component of the input to the neural network model in order to understand what information is more relevant for this task. Table \ref{table:ablation-results} shows the differences with respect to the performance in the validation set. 

The image representation does not seem to be fully exploited by the model. While in Task 2 the image is expected to be superfluous to detect the dialogue act, it should play a role for Task 1, as it imposes constraints on possible actions. It is possible that the off-the-shelf pretrained model is not adequate to encode cliparts and further investigation with other models and fine-tuning is required.

The last message is the most relevant signal for Task 2, as expected, given that it is the iCR being classified. Without it, the task is almost equivalent to Task 1 and the performance is indeed similar. Interestingly, the most relevant signal for Task 1 is the context and not the last utterance, which is evidence that the model fails to distinguish well which instructions require an iCR. To further investigate this, we remove the teller's utterances and the drawer's utterances from the context embeddings. While removing the teller's utterances causes little change, removing the drawer's utterances is almost as detrimental as removing the whole context. We thus conclude that the model is likely exploring patterns in the drawer's behavior to make decisions.\\

\begin{table}[h!]
    \centering
    \small
    \begin{tabular}{r c c c c} 
    \toprule
              &     \multicolumn{2}{c}{\textbf{Task 1: \textit{IF}}} &  \multicolumn{2}{c}{\textbf{Task 2: \textit{IG}}} \\
    \cmidrule(lr){2-3} \cmidrule(lr){4-5}
       &  AP    & mF1 & AP    & mF1 \\
     \cmidrule{1-5} 

        no image & -.032 & -.012 & ~.001 & ~.005 \\
        no message & -.050 & -.021 & -.652 & -.328 \\
        no context & -.109 & -.054 & ~.001 & .007 \\ 
        context w/o teller & -.001 & ~.000 & -.001 & -.000 \\ 
        context w/o drawer & -.087 & -.054 & -.000 & ~.007 \\
                                  
    \bottomrule
    \end{tabular}
    \caption{Results of ablation in the input components. Differences in relation to the main result in the val set.}
    \label{table:ablation-results}
\end{table}

%% file: contents/discussion.tex
Our findings are aligned with the recent conclusions by \citet{aliannejadi-etal-2021-building} and \citet{shi-etal-2022-learning} that the task of predicting when a CR should be made is rather difficult with data-driven models. Techniques to deal with the class imbalance (downsampling, upsampling and varying the cost-sensitive loss function) and variations of the models (\textit{e.g.}\ Transformer-based architectures) so far led us to similar results. On the other hand, the task of identifying iCR utterances is uncomplicated even for a simpler logistic regression model.

The results reached by our model in Task 1 do not quite allow us to see desideratum \textbf{regularity} as satisfied at this point, but we are confident that there is much room for interesting further research with this dataset. On their own, these tasks model an overhearer that predicts what the agent should do. What is of interest in reality is having them integrated as subcomponents, implicitly or explicitly, in the models that also make the instruction-giving/following decisions, because these capabilities are not detached in the agents \textit{de facto}. We expect that the decision to ask for clarification should emerge more easily in representations of models that are also making actions. 

The fact that the drawer's utterances seem to be informative in the dialogue representations for the task speaks against the ``silent drawer assumption'' int he original models \cite{kim-etal-2019-codraw}. Removing the drawer's utterances from the dialogue likely cause loss of relevant dialogue phenomena that is pertinent to the game.

%% file: contents/conclusion.tex
We have shown that CoDraw-iCR (v1), the CoDraw dataset augmented with our iCR annotation, is a valuable resource for investigating instruction-level CRs at scale. Through the corpus analysis, we have also concluded that iCR turns and post-iCR turns imply different game dynamics, which is relevant for models trained to play this game successfully. Therefore, in order to succeed in this type of task, agents need to know how to handle iCRs, as they influence not only the dialogue acts but also the game moves.

Our models perform well on detecting iCRs and lay the groundwork for further research on predicting when an iCR should be made. The research roadmap is to integrate iCRs into the full $IF$ agent, so that the decision to ask for clarification is learnt together with the actions in the game.

The second annotation phase will provide fine-grained categories of iCRs' form and content and ground them to the game objects, opening the possibility to explore other tasks like generation.

%% file: contents/limitations.tex
In this section, we discuss some limitations that we inherit from the CoDraw dataset, and then some limitations of our task setup and baseline model.

CoDraw is a simplified but representative instance of instruction giving/following dialogue games and we show that iCRs are frequent and play an important role in it. Since modelling CRs is still an open problem, using abstract scenes is a reasonable strategy to simplify the underlying task while still giving room for iCRs to occur. Limitations are inherent to data collections in controlled environments. We aim for our annotations to add to other recent efforts, which are limited in other ways. CoDraw-iCR (v1) thus aims to move one step forward towards modelling iCRs, but general conclusions depend on various resources and further collaborative efforts in our field. 

Actions were not irreversible in CoDraw games. The introduction of the peek action for the teller can be an incentive both for the teller to not give exhaustive instructions and for the drawer to build only an approximation, knowing it could be refined after the peek. We have no access to what the performance would have been if they could not make CRs at all.

Meta-data about crowdworker ID is not available.\footnote{Personal communication with the authors.} Because of that, we cannot investigate the effects of individual CR strategies by players. Players that play multiple games get to know what to expect of the game and should both have more practice in identifying underspecified instructions that require repair and be able to make better guesses about the cliparts. Experienced tellers probably anticipate common problems and adapt their instructions to avoid them (\textit{e.g.}\ they know that multiple cliparts of trees exist and would likely describe it in their instruction, avoiding unnecessary communication problems). Besides, we cannot draw conclusions on whether dialogues without iCRs indeed did not require repair or some players were personally less inclined to make the effort to ask for clarification.

Although CRs annotation should take into account the full context \citep{benotti-blackburn-2021-recipe}, the decision to annotate utterance types instead of full dialogues, as discussed in Section \ref{sec:data}, is due to the limited resources given the size of the dataset and to the nature of the game setting. We avoided the need to go over multiple non-iCR utterances that occur very often. The plan for the second step of the annotation is to provide fine-grained annotation for each identified iCR within its own context.

Our models do not take into account the gallery of cliparts available to the drawer, which is informative (as it limits the choices of cliparts per game) and could be part of the input. Preliminary experiments did not lead to better results. Building a suitable representation of the gallery is left for future research.

%% file: contents/appendix-data-statement.tex
\begin{figure*}
  \includegraphics[trim={0cm 5cm 0cm 5cm},clip,width=\linewidth]{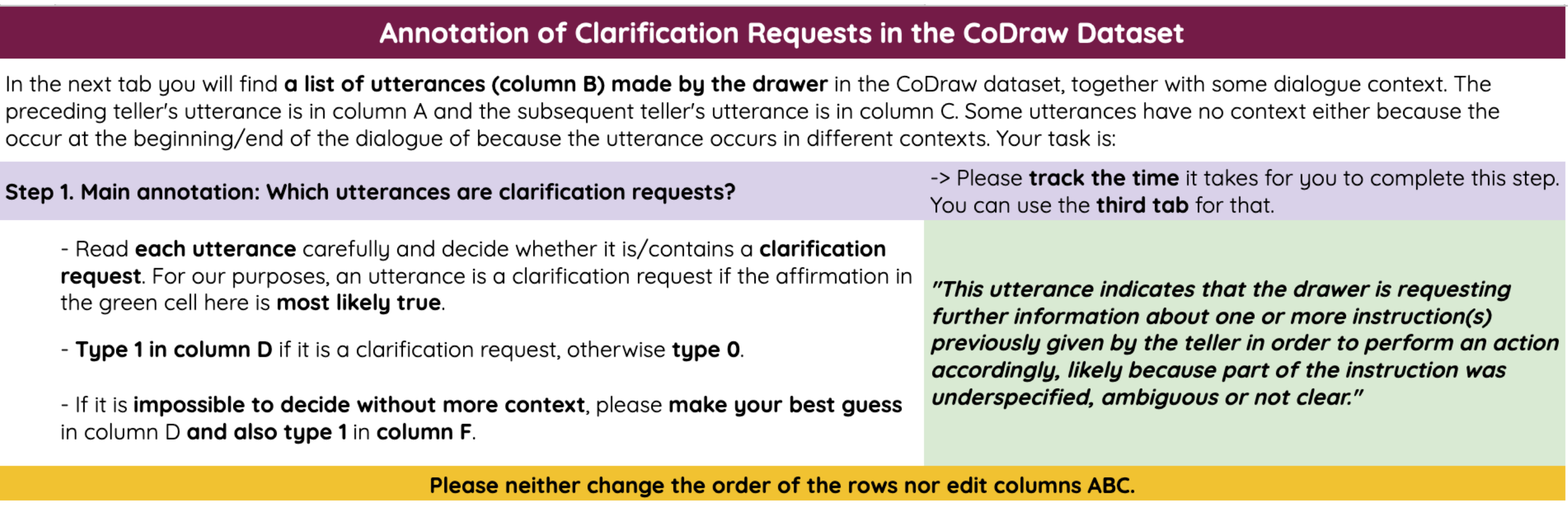}
  \caption{Instructions for the first annotator.}
  \label{fig:instructions}
\end{figure*}

Following \citet{bender-friedman-2018-data}, in this section we provide information about the extended dataset. Figure \ref{fig:instructions} shows the instructions given to the first annotator.\footnote{The second annotator got more detailed instructions to perform the fine-grained annotation, which is not part of this publication. For the analysis and experiments in this work, we use the labels by the second annotator, who went through a more extensive background reading about CRs. There are few disagreements, as shown in the main text.} 

\textbf{Curation Rationale}. We annotate iCRs in all dialogues of the CoDraw dataset \cite{kim-etal-2019-codraw}, which contains 9,993 dialogues produced by crowdworkers and has been released under a \href{https://creativecommons.org/licenses/by-nc/4.0/}{CC BY-NC 4.0} license. Please refer to the original paper for details about their data collection.

\textbf{Language Variety and Speaker Demographic}. The CoDraw dataset comprises written interaction in English, however no information about crowd worker demographic has been released in the dataset repository. 

\textbf{Annotator Demographic}. The annotators who identified iCRs in CoDraw are a male and a female Computational Linguistics bachelor students who are non-native fluent English speakers working at colabPotsdam as student assistants. The students were paid according to the German’s regulation for student assistants.

\textbf{Situation}. In CoDraw, crowdworkers exchanged written messages of up to 140 characters via a chat interface in a crowdsourcing tool. \textit{IF} and \textit{IG} had to send messages in alternating turns. The interaction was synchronous and task-oriented. 

\textbf{Text Characteristics}. The CoDraw authors pre-process all collected utterances using a spell checker and tokenize the text with a natural language toolkit.

%% file: contents/appendix-annotation.tex
We provide here further descriptive characteristics of the annotated dataset.

Table \ref{table:cr-examples} shows a few negative examples, \textit{i.e.} utterances that are not iCRs. Although utterances like \textit{what do you see in the sky ?} and \textit{anything else to change ?} can indeed be considered task-level CRs, we do not consider them iCRs because they do not directly refer back to a given instruction.

    \begin{table}[h!]
       \centering
       \small
      \begin{tabular}{l l} 
         \toprule
            & yeah it was a lot but thanks for finishing \\
            & i am a patient worker ready to start \\
            & check please and tell me what to change \\
            & anything else to change ? \\
            & what else is in the picture and where ? \\
            & i 've made all the changes you 've listed . \\
            & ok and look \\
            & alright , made changes \\
            & please be more specific thanks \\
            & ok anything else in the picture ? \\
            & yes , please lmk of any corrections \\
            & ok i got that : \\
            & ready whenever you are . : \\
            & what what is the first object and location ? \\
            & alright done \\
            & what do you see in the sky ? \\
            & tell me what we have \\
        \bottomrule
        \end{tabular}
        \caption{Negative examples in CoDraw.}
        \label{table:cr-examples}
    \end{table}
    
\begin{figure}
  \includegraphics[trim={0cm 0cm 0cm 0cm},clip,width=\linewidth]{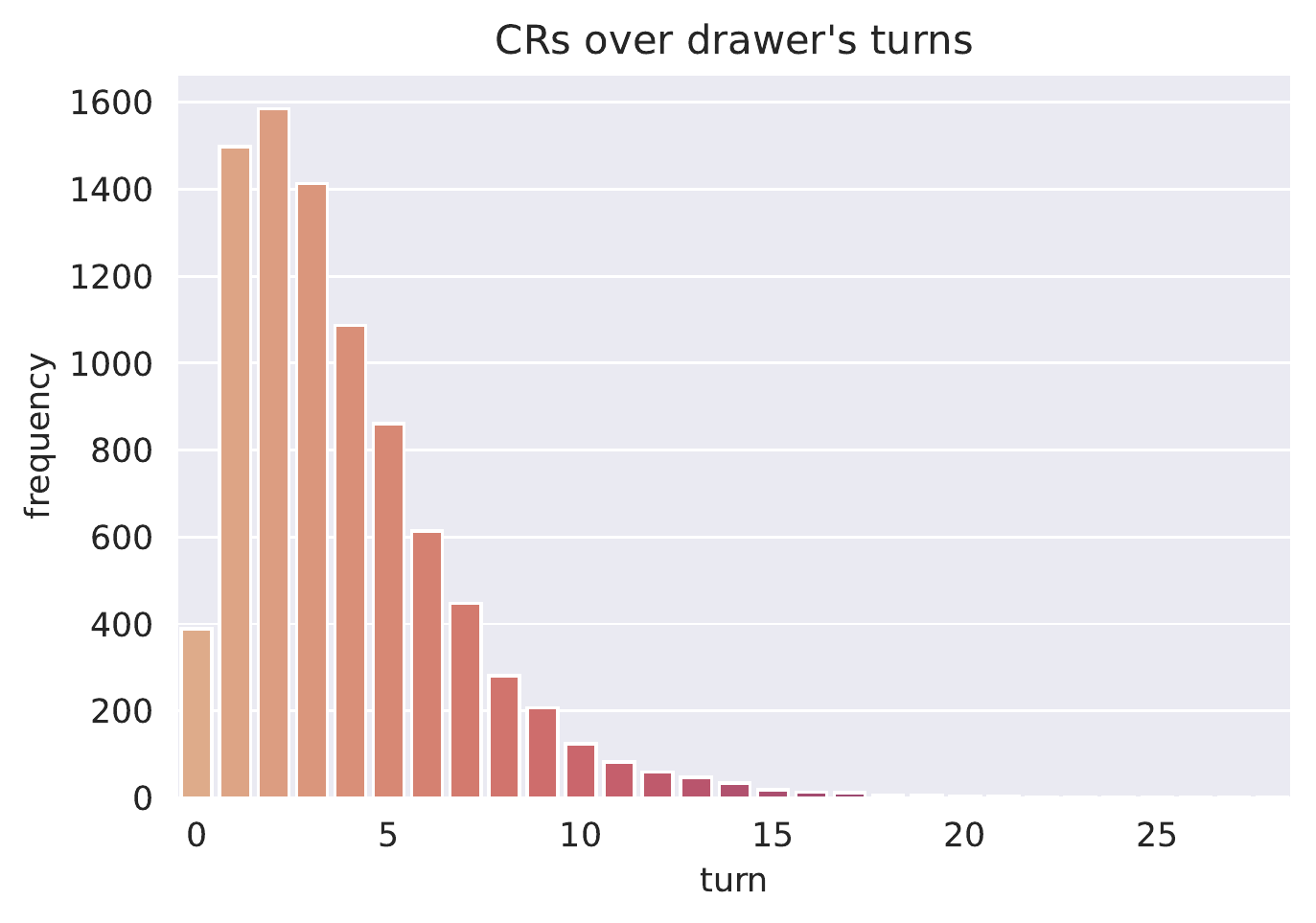}
  \caption{In which round iCRs occur.}
  \label{fig:which-round}
\end{figure}

\begin{figure}
	\centering
	\includegraphics[trim={0cm 0cm 0cm 0cm},clip,width=5cm]{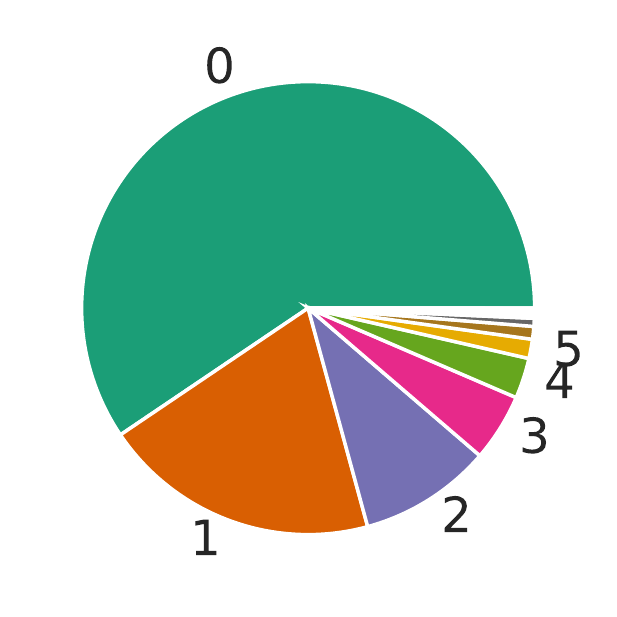}
	\caption{Number of iCRs per dialogue.}
	\label{fig:n-cr-dialogue}
\end{figure}

Figure \ref{fig:which-round} depicts in which rounds iCRs occur in the corpus. Given that the average dialogue length is 7.7 rounds, most instruction CRs in this dataset are occurring early in their corresponding dialogue. The distribution of the number of iCRs per dialogue is illustrated in Figure \ref{fig:n-cr-dialogue}, where we see that it is very rare to have dialogues with more than 5 iCRs. 

\begin{figure}
	\includegraphics[trim={0cm 0cm 0cm 0cm},clip,width=\linewidth]{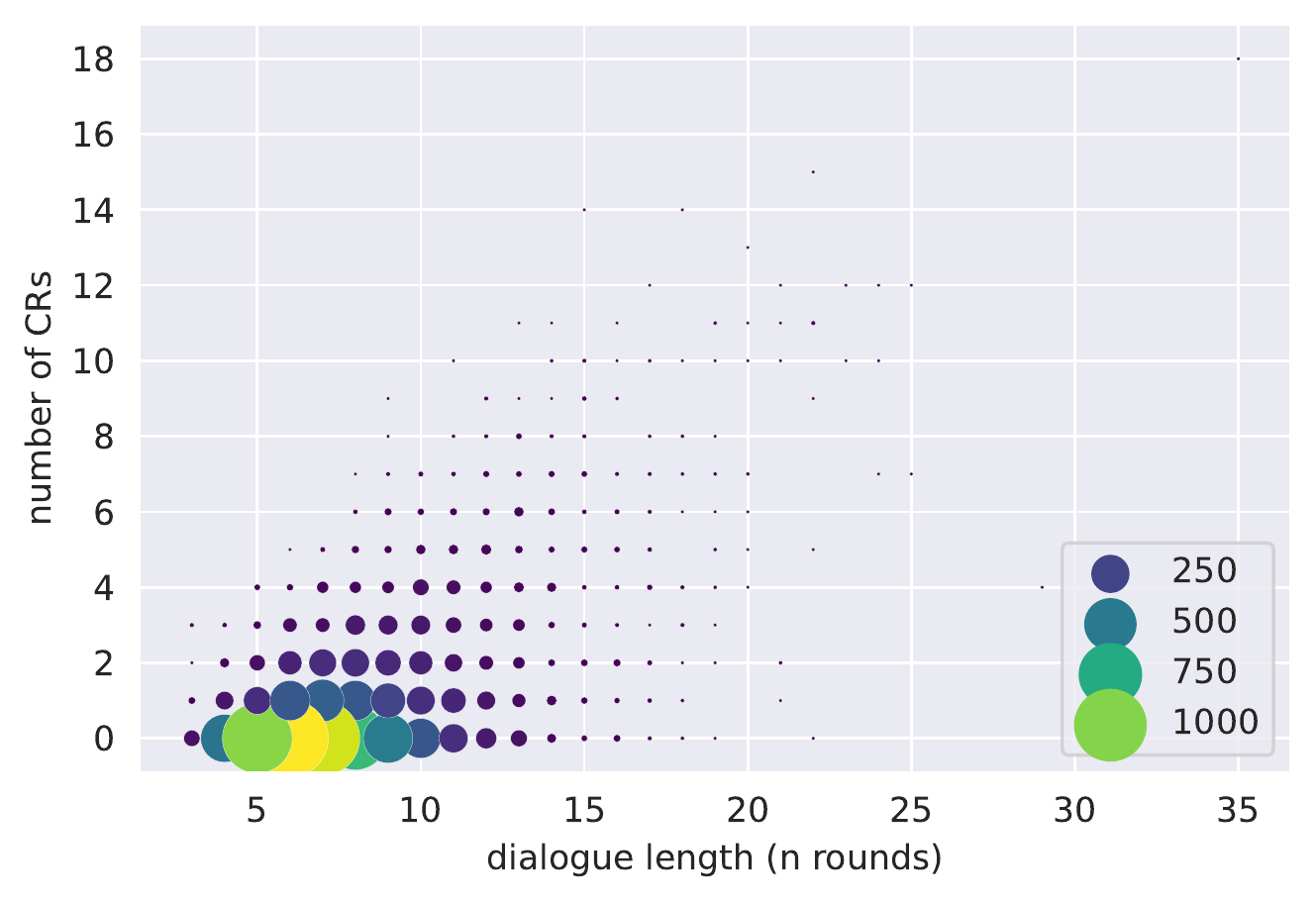}
	\caption{Number of iCRs vs. number of rounds.}
	\label{fig:n-cr-n-rounds}
\end{figure}

Figure \ref{fig:n-cr-n-rounds} breaks down the number of iCRs per number of dialogue rounds. Dialogues with more than 10 iCRs are outliers, which is expected given that most dialogues have no more then 15 rounds.

14.58\% of validation and 11.96\% of test iCR utterance types also occur in the training set. 17.50\% of validation and 14.81\% of test iCR utterances also occur in the training set. The overlap is low, which is a desirable characteristic to reduce the memorization shortcuts for models trained on this dataset. \\

\noindent \textbf{Computing actions and score differences}. We group the drawing actions into three main categories: \textit{addition} (when a clipart is added to the scene), \textit{edit} (when some change occurs with an object that existed in the previous round), and \textit{no action} if the drawer did not perform an action in a round. Edits can be deletion, move (position change), flip and resize. We do not track whether newly added cliparts get immediately flipped or resized in relation to the gallery when they are added. We noticed that some actions that visually seem to be only flips or resize happen together with a move. However, our inspection shows that this does not occur consistently with a specific subset of cliparts and there is also a portion of cases where flips and resizes occur without moves. Therefore, each move, resize, flip are counted as one separate action in our analysis. We compute the moves based on the differences over consecutive rounds in sequence of scene strings labeled \textit{abs\_d} provided in the original dataset, assuming that they describe the state of the canvas at the moment when the drawer sends their message, \textit{i.e.} at the end of the current round.

Scene similarity is computed with the scripts made available by the CoDraw authors on \href{https://github.com/facebookresearch/codraw-models}{GitHub}.

%% file: contents/appendix-reproducibility.tex
In this section we describe the details of the implementation and datasets for reproducibility purposes. Further information and documentation is available in the code repository.

The random and trivial baselines are trained with \href{https://scikit-learn.org/stable/index.html}{scikit-learn} (v1.1.2) with class weight set to balanced. A maximum of 1,000 iterations was still not sufficient for convergence is all cases. The hypothesis tests are carried out with \href{https://scipy.org/}{SciPy} (v1.10.0), using the permutation test for the difference of means, with type set to independent.\\

\noindent \textbf{Models}. Our models are implemented with \href{https://pytorch.org/}{PyTorch} (v1.11.0) and \href{https://www.pytorchlightning.ai/}{PyTorch Lightning} (v1.6.4). Metrics are computed using \href{https://torchmetrics.readthedocs.io/en/stable/}{TorchMetrics} (v0.10.0). The experiments were run in Linux 5.4.0-99-generic, machine/processor x86\_64 in Python 3.9.12 on an NVIDIA GeForce GTX 1080 Ti GPUs with CUDA v11.6. The architecture of the full neural network model with its corresponding layers and dimensions were:

\begin{small}
\begin{verbatim}

  (model): Classifier(
    (img_encoder): ImageEncoder(
      (encoder): Linear(in_features=2048, 
                        out_features=128, 
                        bias=True)
    )
    (msg_encoder): TextEncoder(
      (encoder): Linear(in_features=768, 
                        out_features=128, 
                        bias=True)
    )
    (context_encoder): TextEncoder(
      (encoder): Linear(in_features=768, 
                        out_features=128, 
                        bias=True)
    )
    (classifier): DeeperClassifier(
      (classifier): Sequential(
        (0): LeakyReLU(negative_slope=0.01)
        (1): Dropout(p=0.1, inplace=False)
        (2): Linear(in_features=384, 
                    out_features=256, 
                    bias=True)
        (3): BatchNorm1d(256, 
                         eps=1e-05, 
                         momentum=0.1,
                         affine=True,
                              track_running_stats=True)
        (4): LeakyReLU(negative_slope=0.01)
        (5): Linear(in_features=256, 
                    out_features=1, 
                    bias=True)
      )
    )
  )

\end{verbatim}
\end{small}

The complete model has 558,465 trainable parameters. For the ablation experiments, the number of dimensions was reduced according to the input, and the number of parameters were 263,425 (no image), 427,265 (no context) and 427,265 (no utterance).

Training is carried out with the Adam optimizer \citep{kingma2014adam} to minimize weighted binary cross entropy (the weight is the hyperparameter weight cr) estimated with a sigmoid function applied to the output logits. We use the Bayes algorithm from \href{comet.ml}{comet.ml} to perform hyperparameter search seeking to maximize Average Precision (and also AUC of the Precision-Recall curve in some preliminary experiments) on the validation set for Task 1. For the final version, we run 111 experiments during hyperparameter search. The optimal hyperparameters used in the experiments are shown in Table \ref{table:hyperparameters} together with their corresponding bounds. We use the second best performing configuration, because it is only around 6e-7 below the best one, but has more stable learning curves. All other experiments (ablation and Task 2) use the same configuration, except that the dimensions change according to the input vectors for ablation. We report the results of one run using the best configuration.


        \begin{table*}[h!]
            \centering
            \small
            \begin{tabular}{p{3.1cm} l l l} 
             \toprule
              \textbf{hyperparameter} & \textbf{type} & \textbf{search bounds} & \textbf{optimal value} \\ 
             \midrule
                accumulate gradient & discrete & 1, 2, 5, 10, 25 & 25 \\
                batch size & discrete & 32, 64, 128, 256, 512, 1024 & 128 \\
                clipping  & discrete & 0, 0.25, 0.5, 1, 2.5, 5, 10 & 1 \\
                dropout & discrete & 0.1, 0.2, 0.3, 0.5 & 0.1 \\
                gamma & discrete & 0.1, 0.5, 0.9, 0.99, 1 & 0.99 \\
                hidden dimension & discrete & 32, 64, 128, 256, 512, 1024 & 256 \\
                internal embeddings dim & discrete & 32, 64, 128, 256, 512, 1024 & 128 \\
                learning rate & discrete & 0.1, 0.01, 0.001, 0.0001, 0.003, 0.0003, 0.00001, 0.0005 & 0.003 \\
                lr scheduler & categorical & none, exp, step & exp \\
                lr step & integer & min=1, max=5 & 2 \\
                random seed & integer & min=1, max=54321 & 35466 \\
                weight cr & float & min=1, max=10 & 2.6125454767515217 \\
                weight decay & discrete & 1, 0.1, 0.01, 0.001, 0.0001 & 0.0001\\
            \bottomrule
            \end{tabular}
            \caption{Hyperparameters: Search bounds and optimal values.}
            \label{table:hyperparameters}
        \end{table*}

 We use a decision threshold of 0.5 for the evaluation metrics that require a fixed threshold.

We train the models for up to 20 epochs, which takes around 3-4 minutes, including inference, which requires around 4 seconds. Although it takes more than 20 epochs to achieve a higher performance on the training set, the maximum for the validation set is reached in early epochs. We then use the model checkpoint with the highest Average Precision on the validation set to run evaluation on the test set.

We use the \texttt{all-mpnet-base-v2} model from SentenceTransformers to encode the texts into a representation with 768 dimensions. The image representation has 2048 dimensions. Images are preprocessed according to \href{https://github.com/pytorch/examples/blob/97304e232807082c2e7b54c597615dc0ad8f6173/imagenet/main.py#L197-L198}{PyTorch Vision} documentation (without resizing and centering) and features are extracted following \href{https://discuss.pytorch.org/t/how-can-l-use-the-pre-trained-resnet-to-extract-feautres-from-my-own-dataset/9008/6}{recommendation} on their forum. We use the pretrained model \texttt{resnet101} available from \href{https://pytorch.org/vision/0.12/}{torchvision} (v0.12.0).\\

\noindent \textbf{Datasets}. We use the same train/val/test splits as the original CoDraw dataset. The sizes and the distribution of labels in the annotated dataset is in Table \ref{table:icr-labels}. For retrieving the context embeddings, we add a /T token before the teller's utterance and a /D token before the drawer's utterances. We also add a /PEEK token before the utterances of the round when a peek action occurs. The context is limited to the last 200 tokens. Utterances are tokenized with blank spaces on the preprocessed published dataset.



%% file: contents/appendix-results.tex
We present more details about the performance on the validation set. 

    \begin{figure}[ht]
    \centering
      \includegraphics[trim={0cm 0cm 0cm 0cm},clip,width=7cm]{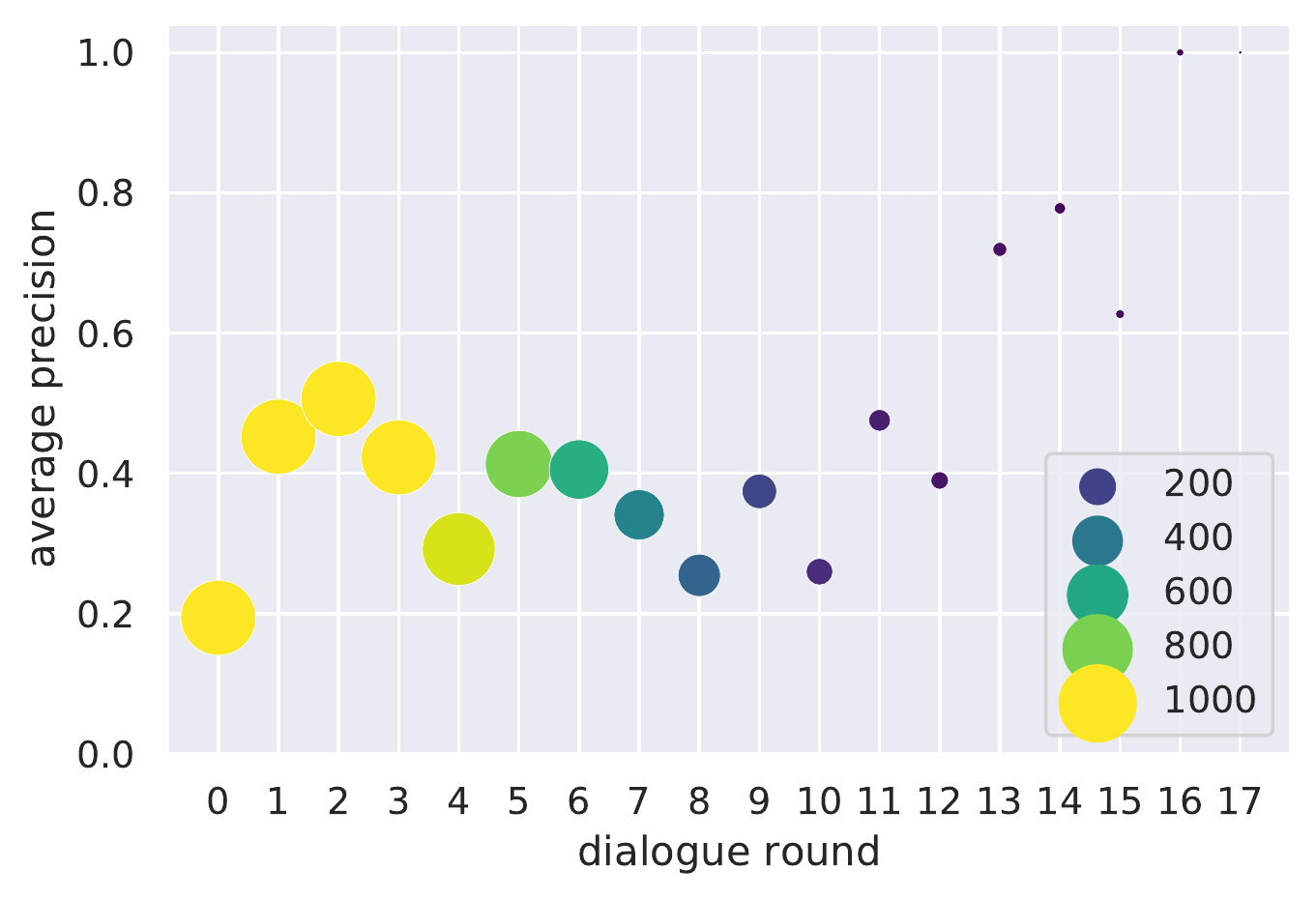}
      \caption{Average precision per round (validation set).}
      \label{fig:avp-per-round}
    \end{figure}

\begin{figure}[ht]

        \begin{subfigure}{\columnwidth}
             \centering
             \includegraphics[trim={0cm 0cm 0cm 0cm},clip,width=\columnwidth]{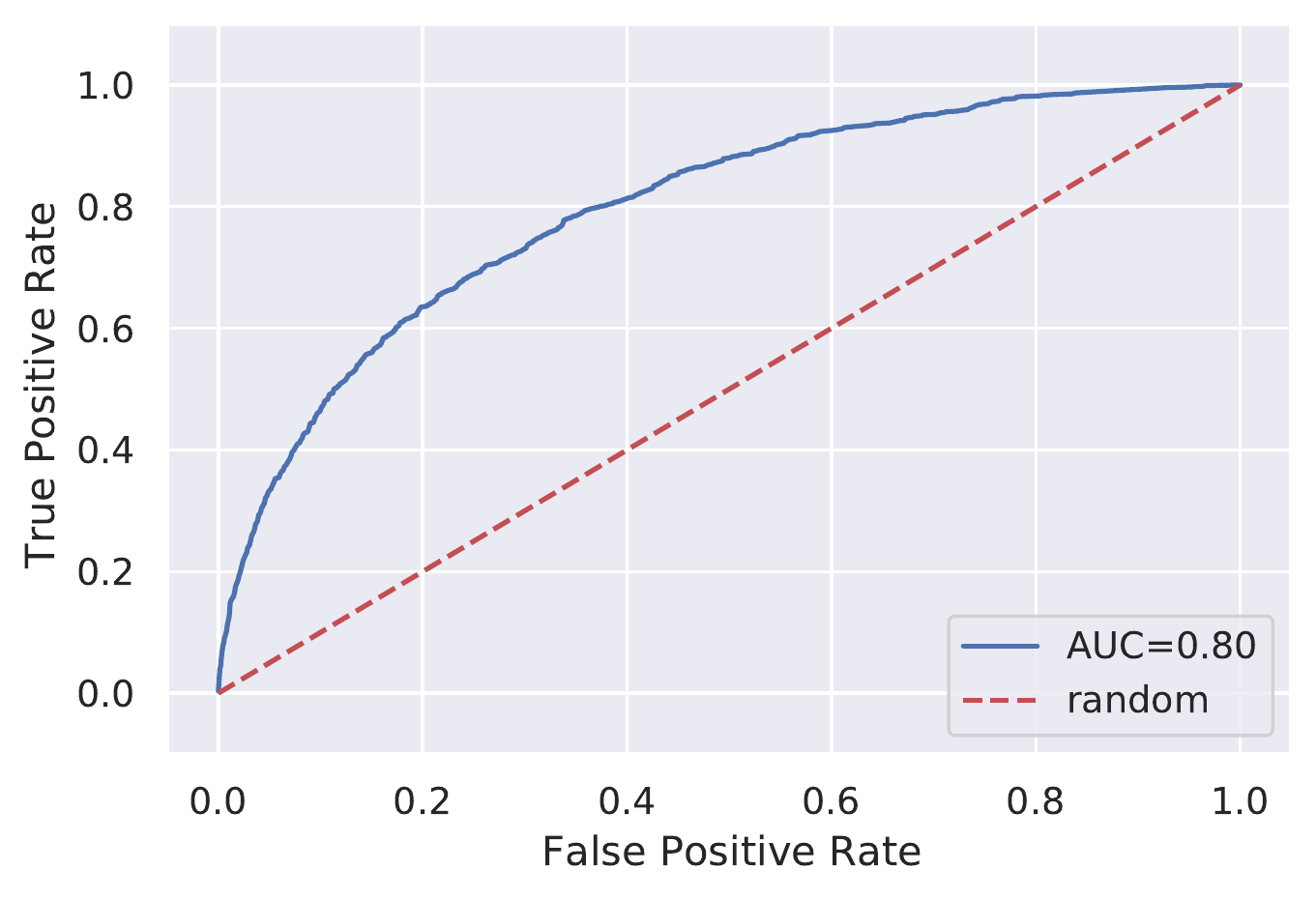}%
             \caption{Task 1: \textit{IF}}
         \end{subfigure}

        \begin{subfigure}{\columnwidth}
             \centering
             \includegraphics[trim={0cm 0cm 0cm 0cm},clip,width=\columnwidth]{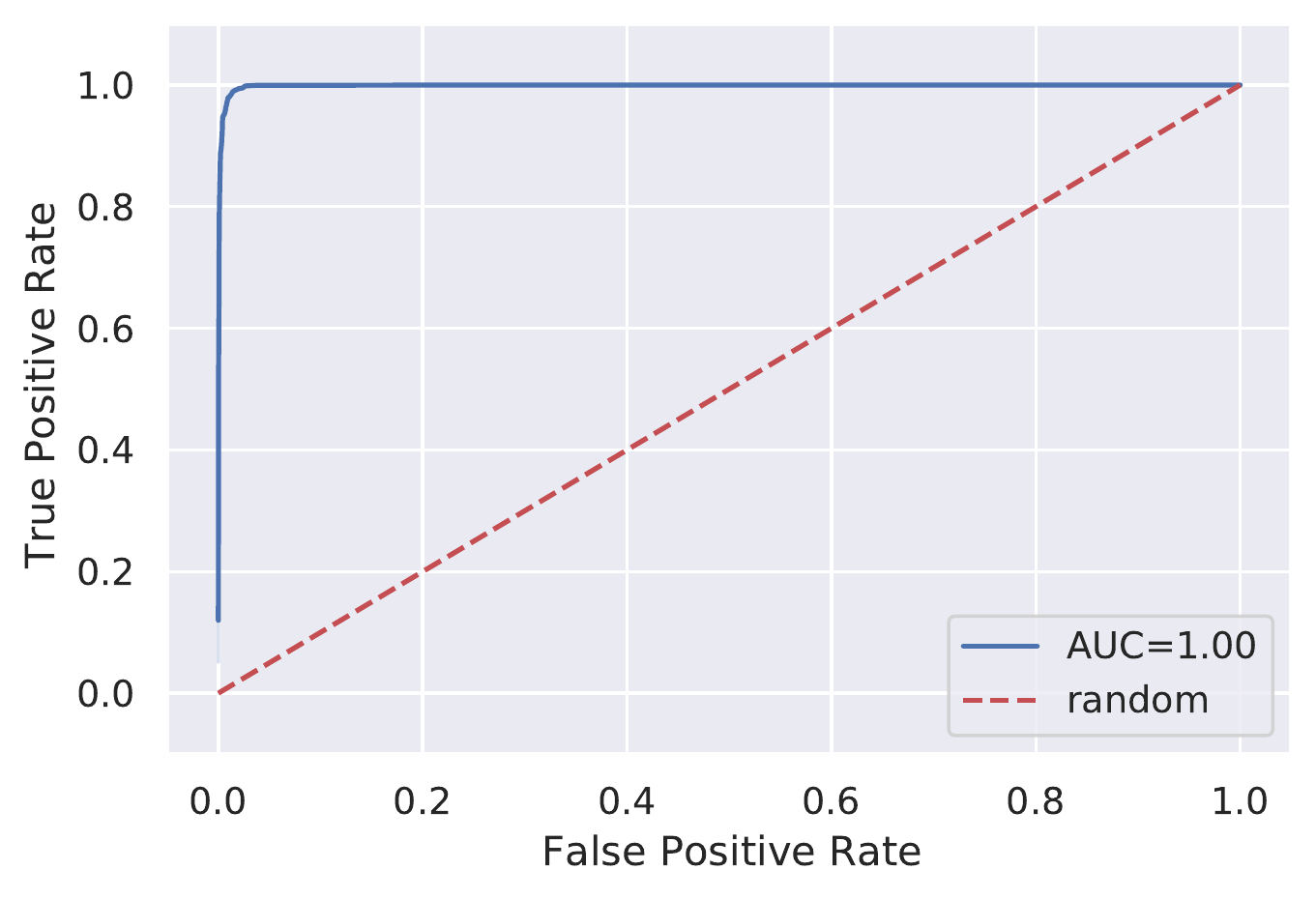}%
             \caption{Task 2: \textit{IG}}
         \end{subfigure}
        \caption{ROC curves (validation set).}
        \label{fig:roc}
    
    \end{figure}

\begin{figure}[ht]

        \begin{subfigure}{\columnwidth}
             \centering
             \includegraphics[trim={0cm 0cm 0cm 0cm},clip,width=\columnwidth]{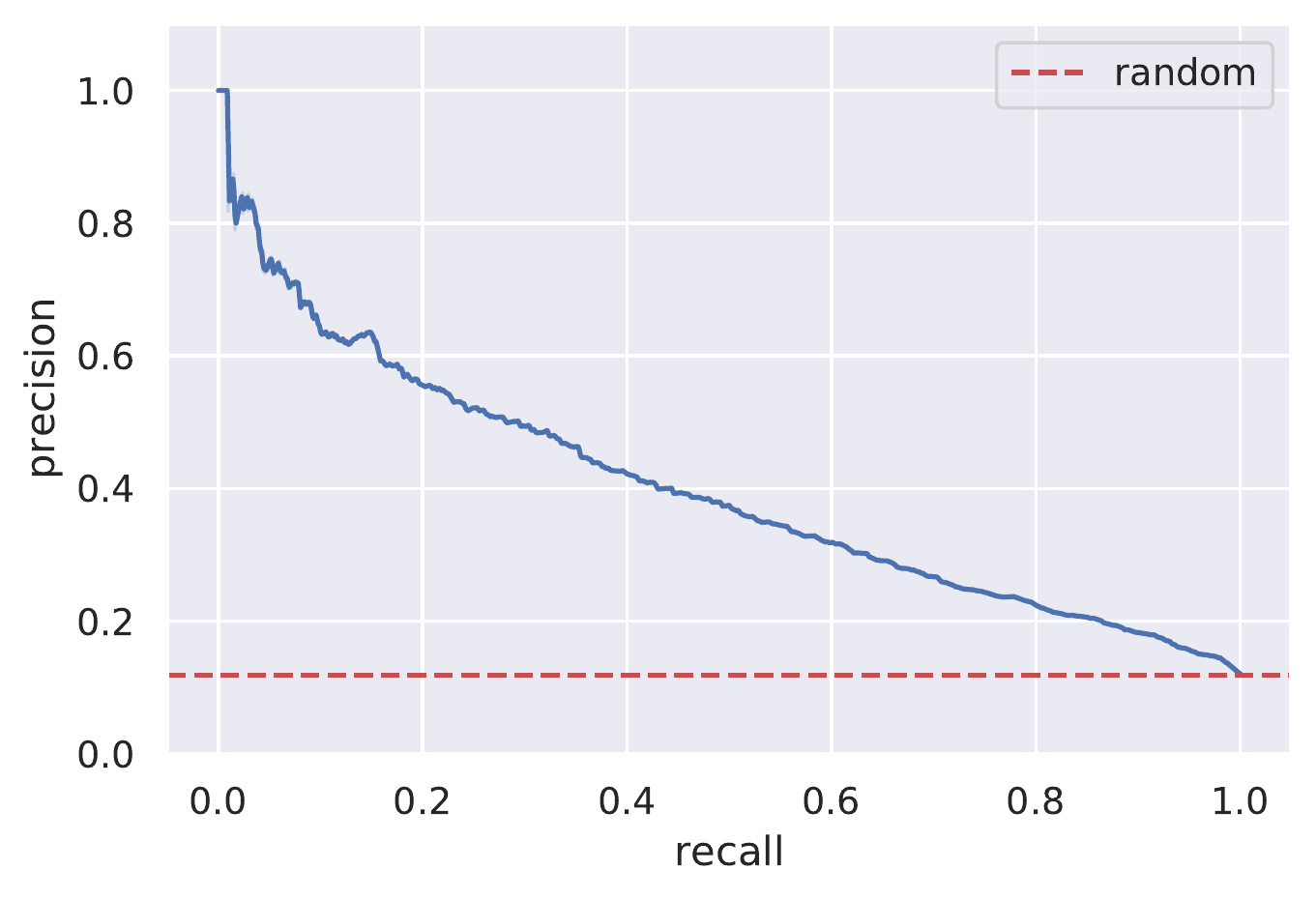}%
             \caption{Task 1: \textit{IF}}
         \end{subfigure}

        \begin{subfigure}{\columnwidth}
             \centering
             \includegraphics[trim={0cm 0cm 0cm 0cm},clip,width=\columnwidth]{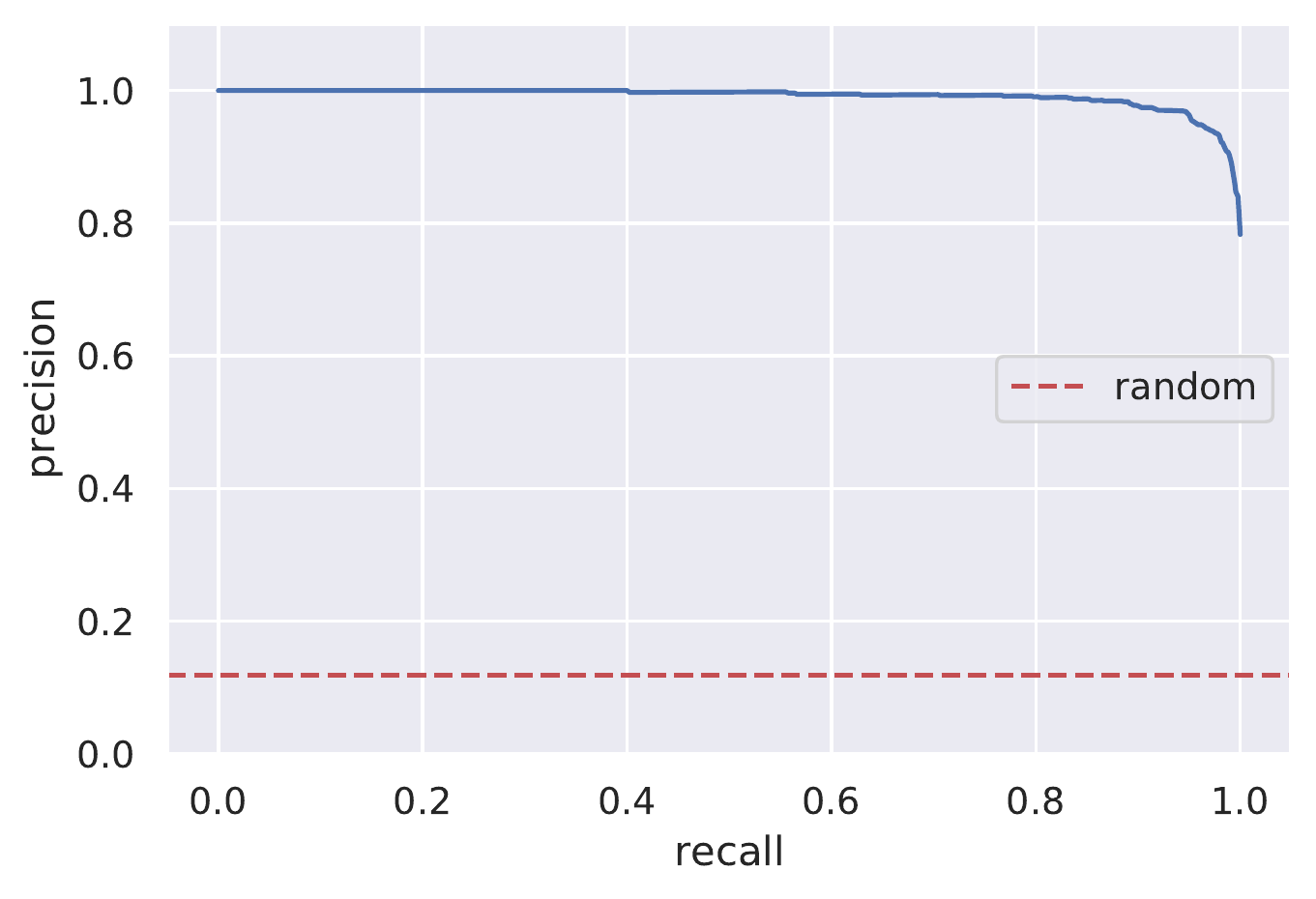}%
             \caption{Task 2: \textit{IG}}
         \end{subfigure}

        \caption{Precision-Recall curves (validation set).}
        \label{fig:precrec}
    \end{figure}

Figure \ref{fig:avp-per-round} shows the AP metric split by round, Figure \ref{fig:roc} presents the ROC curves and Figure \ref{fig:precrec}, the Precision Recall curves of the best checkpoint.

\vspace{0.9cm}

%% file: main.bbl
\begin{thebibliography}{64}
\expandafter\ifx\csname natexlab\endcsname\relax\def\natexlab#1{#1}\fi

\bibitem[{Aliannejadi et~al.(2021)Aliannejadi, Kiseleva, Chuklin, Dalton, and
  Burtsev}]{aliannejadi-etal-2021-building}
Mohammad Aliannejadi, Julia Kiseleva, Aleksandr Chuklin, Jeff Dalton, and
  Mikhail Burtsev. 2021.
\newblock \href {https://doi.org/10.18653/v1/2021.emnlp-main.367} {Building and
  evaluating open-domain dialogue corpora with clarifying questions}.
\newblock In \emph{Proceedings of the 2021 Conference on Empirical Methods in
  Natural Language Processing}, pages 4473--4484, Online and Punta Cana,
  Dominican Republic. Association for Computational Linguistics.

\bibitem[{Aliannejadi et~al.(2019)Aliannejadi, Zamani, Crestani, and
  Croft}]{aliannejadi2019asking}
Mohammad Aliannejadi, Hamed Zamani, Fabio Crestani, and W~Bruce Croft. 2019.
\newblock Asking clarifying questions in open-domain information-seeking
  conversations.
\newblock In \emph{Proceedings of the 42nd international acm sigir conference
  on research and development in information retrieval}, pages 475--484.

\bibitem[{Allwood(2000)}]{allwood2000activity}
Jens Allwood. 2000.
\newblock An activity-based approach to pragmatics.
\newblock \emph{Gothenburg Papers in Theorecial Linguistics}, 76.

\bibitem[{Bender and Friedman(2018)}]{bender-friedman-2018-data}
Emily~M. Bender and Batya Friedman. 2018.
\newblock \href {https://doi.org/10.1162/tacl_a_00041} {Data statements for
  natural language processing: Toward mitigating system bias and enabling
  better science}.
\newblock \emph{Transactions of the Association for Computational Linguistics},
  6:587--604.

\bibitem[{Benotti(2009)}]{benotti-2009-clarification}
Luciana Benotti. 2009.
\newblock \href {https://aclanthology.org/W09-3929} {Clarification potential of
  instructions}.
\newblock In \emph{Proceedings of the {SIGDIAL} 2009 Conference}, pages
  196--205, London, UK. Association for Computational Linguistics.

\bibitem[{Benotti and Blackburn(2017)}]{benotti2017modeling}
Luciana Benotti and Patrick Blackburn. 2017.
\newblock Modeling the clarification potential of instructions: Predicting
  clarification requests and other reactions.
\newblock \emph{Computer Speech \& Language}, 45:536--551.

\bibitem[{Benotti and Blackburn(2021)}]{benotti-blackburn-2021-recipe}
Luciana Benotti and Patrick Blackburn. 2021.
\newblock \href {https://doi.org/10.18653/v1/2021.naacl-main.320} {A recipe for
  annotating grounded clarifications}.
\newblock In \emph{Proceedings of the 2021 Conference of the North American
  Chapter of the Association for Computational Linguistics: Human Language
  Technologies}, pages 4065--4077, Online. Association for Computational
  Linguistics.

\bibitem[{Bohus and Rudnicky(2005)}]{bohus-rudnicky-2005-sorry}
Dan Bohus and Alexander~I. Rudnicky. 2005.
\newblock \href {https://aclanthology.org/2005.sigdial-1.14} {Sorry and {I}
  didn{'}t catch that! - an investigation of non-understanding errors and
  recovery strategies}.
\newblock In \emph{Proceedings of the 6th SIGdial Workshop on Discourse and
  Dialogue}, pages 128--143, Lisbon, Portugal. Special Interest Group on
  Discourse and Dialogue (SIGdial).

\bibitem[{Braslavski et~al.(2017)Braslavski, Savenkov, Agichtein, and
  Dubatovka}]{braslavski2017cqa}
Pavel Braslavski, Denis Savenkov, Eugene Agichtein, and Alina Dubatovka. 2017.
\newblock \href {https://doi.org/10.1145/3020165.3022149} {What do you mean
  exactly? analyzing clarification questions in cqa}.
\newblock In \emph{Proceedings of the 2017 Conference on Conference Human
  Information Interaction and Retrieval}, CHIIR '17, page 345–348, New York,
  NY, USA. Association for Computing Machinery.

\bibitem[{Clark(1996)}]{clark1996using}
Herbert~H Clark. 1996.
\newblock \emph{Using language}.
\newblock Cambridge university press.

\bibitem[{Cohen(1960)}]{cohen-kappa}
Jacob Cohen. 1960.
\newblock \href {https://doi.org/10.1177/001316446002000104} {A coefficient of
  agreement for nominal scales}.
\newblock \emph{Educational and Psychological Measurement}, 20(1):37--46.

\bibitem[{De~Boni and Manandhar(2003)}]{de-boni-manandhar-2003-analysis}
Marco De~Boni and Suresh Manandhar. 2003.
\newblock \href {https://aclanthology.org/N03-1007} {An analysis of
  clarification dialogue for question answering}.
\newblock In \emph{Proceedings of the 2003 Human Language Technology Conference
  of the North {A}merican Chapter of the Association for Computational
  Linguistics}, pages 48--55.

\bibitem[{Deits et~al.(2013)Deits, Tellex, Thaker, Simeonov, Kollar, and
  Roy}]{deits-2013-clarifying}
Robin Deits, Stefanie Tellex, Pratiksha Thaker, Dimitar Simeonov, Thomas
  Kollar, and Nicholas Roy. 2013.
\newblock \href {https://doi.org/10.5898/JHRI.2.2.Deits} {Clarifying commands
  with information-theoretic human-robot dialog}.
\newblock \emph{J. Hum.-Robot Interact.}, 2(2):58–79.

\bibitem[{El-Nouby et~al.(2019)El-Nouby, Sharma, Schulz, Hjelm, Asri, Kahou,
  Bengio, and W.Taylor}]{ElNouby2019TellDA}
Alaaeldin El-Nouby, Shikhar Sharma, Hannes Schulz, Devon Hjelm, Layla~El Asri,
  Samira~Ebrahimi Kahou, Yoshua Bengio, and Graham W.Taylor. 2019.
\newblock Tell, draw, and repeat: Generating and modifying images based on
  continual linguistic instruction.
\newblock \emph{2019 IEEE/CVF International Conference on Computer Vision
  (ICCV)}, pages 10303--10311.

\bibitem[{Fu et~al.(2020)Fu, Wang, Grafton, Eckstein, and
  Wang}]{Fu2020IterativeLI}
Tsu-Jui Fu, Xin Wang, Scott~T. Grafton, Miguel~P. Eckstein, and William~Yang
  Wang. 2020.
\newblock Iterative language-based image editing via self-supervised
  counterfactual reasoning.
\newblock In \emph{EMNLP}.

\bibitem[{Gabsdil(2003)}]{gabsdil2003clarification}
Malte Gabsdil. 2003.
\newblock Clarification in spoken dialogue systems.
\newblock In \emph{Proceedings of the 2003 AAAI Spring Symposium. Workshop on
  Natural Language Generation in Spoken and Written Dialogue}, pages 28--35.

\bibitem[{Gao et~al.(2022)Gao, Gao, Gong, Lin, Thattai, and
  Sukhatme}]{gao2022dialfred}
Xiaofeng Gao, Qiaozi Gao, Ran Gong, Kaixiang Lin, Govind Thattai, and Gaurav~S
  Sukhatme. 2022.
\newblock Dialfred: Dialogue-enabled agents for embodied instruction following.
\newblock \emph{IEEE Robotics and Automation Letters}, 7(4):10049--10056.

\bibitem[{Gella et~al.(2022)Gella, Padmakumar, Lange, and
  Hakkani-Tur}]{gella-etal-2022-dialog}
Spandana Gella, Aishwarya Padmakumar, Patrick Lange, and Dilek Hakkani-Tur.
  2022.
\newblock \href {https://aclanthology.org/2022.sigdial-1.13} {Dialog acts for
  task driven embodied agents}.
\newblock In \emph{Proceedings of the 23rd Annual Meeting of the Special
  Interest Group on Discourse and Dialogue}, pages 111--123, Edinburgh, UK.
  Association for Computational Linguistics.

\bibitem[{Gervits et~al.(2021)Gervits, Roque, Briggs, Scheutz, and
  Marge}]{gervits-etal-2021-agents}
Felix Gervits, Antonio Roque, Gordon Briggs, Matthias Scheutz, and Matthew
  Marge. 2021.
\newblock \href {https://aclanthology.org/2021.sigdial-1.37} {How should agents
  ask questions for situated learning? an annotated dialogue corpus}.
\newblock In \emph{Proceedings of the 22nd Annual Meeting of the Special
  Interest Group on Discourse and Dialogue}, pages 353--359, Singapore and
  Online. Association for Computational Linguistics.

\bibitem[{Ginzburg(2012)}]{ginzburg_grounding_2012}
Jonathan Ginzburg. 2012.
\newblock \href {https://doi.org/10.1093/acprof:oso/9780199697922.003.0006}
  {Grounding and {CRification}}.
\newblock In \emph{The {Interactive} {Stance}}. Oxford University Press,
  Oxford.

\bibitem[{Ginzburg and Luecking(2021)}]{ginzburg-luecking-2021-requesting}
Jonathan Ginzburg and Andy Luecking. 2021.
\newblock \href {https://aclanthology.org/2021.mmsr-1.3} {Requesting
  clarifications with speech and gestures}.
\newblock In \emph{Proceedings of the 1st Workshop on Multimodal Semantic
  Representations (MMSR)}, pages 21--31, Groningen, Netherlands (Online).
  Association for Computational Linguistics.

\bibitem[{He et~al.(2016)He, Zhang, Ren, and Sun}]{he2016deep}
Kaiming He, Xiangyu Zhang, Shaoqing Ren, and Jian Sun. 2016.
\newblock Deep residual learning for image recognition.
\newblock In \emph{Proceedings of the IEEE conference on computer vision and
  pattern recognition}, pages 770--778.

\bibitem[{Healey et~al.(2003)Healey, Purver, King, Ginzburg, and
  Mills}]{healey2003experimenting}
Patrick~GT Healey, Matthew Purver, James King, Jonathan Ginzburg, and Greg~J
  Mills. 2003.
\newblock Experimenting with clarification in dialogue.
\newblock In \emph{Proceedings of the Annual Meeting of the Cognitive Science
  Society}, volume~25.

\bibitem[{Hu et~al.(2020)Hu, Wen, Wang, Li, and
  de~Melo}]{hu-etal-2020-interactive}
Xiang Hu, Zujie Wen, Yafang Wang, Xiaolong Li, and Gerard de~Melo. 2020.
\newblock \href {https://doi.org/10.18653/v1/2020.coling-industry.8}
  {Interactive question clarification in dialogue via reinforcement learning}.
\newblock In \emph{Proceedings of the 28th International Conference on
  Computational Linguistics: Industry Track}, pages 78--89, Online.
  International Committee on Computational Linguistics.

\bibitem[{Jackson and Williams(2018)}]{jackson2018robot}
Ryan~Blake Jackson and Tom Williams. 2018.
\newblock Robot: Asker of questions and changer of norms.
\newblock \emph{Proceedings of ICRES}.

\bibitem[{Jackson and Williams(2019)}]{jackson2019language}
Ryan~Blake Jackson and Tom Williams. 2019.
\newblock Language-capable robots may inadvertently weaken human moral norms.
\newblock In \emph{2019 14th ACM/IEEE International Conference on Human-Robot
  Interaction (HRI)}, pages 401--410. IEEE.

\bibitem[{Kim et~al.(2019)Kim, Kitaev, Chen, Rohrbach, Zhang, Tian, Batra, and
  Parikh}]{kim-etal-2019-codraw}
Jin-Hwa Kim, Nikita Kitaev, Xinlei Chen, Marcus Rohrbach, Byoung-Tak Zhang,
  Yuandong Tian, Dhruv Batra, and Devi Parikh. 2019.
\newblock \href {https://doi.org/10.18653/v1/P19-1651} {{C}o{D}raw:
  Collaborative drawing as a testbed for grounded goal-driven communication}.
\newblock In \emph{Proceedings of the 57th Annual Meeting of the Association
  for Computational Linguistics}, pages 6495--6513, Florence, Italy.
  Association for Computational Linguistics.

\bibitem[{Kingma and Ba(2015)}]{kingma2014adam}
Diederik~P. Kingma and Jimmy Ba. 2015.
\newblock \href {http://arxiv.org/abs/1412.6980} {Adam: {A} method for
  stochastic optimization}.
\newblock In \emph{3rd International Conference on Learning Representations,
  {ICLR} 2015, San Diego, CA, USA, May 7-9, 2015, Conference Track
  Proceedings}.

\bibitem[{Kiseleva et~al.(2021)Kiseleva, Li, Aliannejadi, Mohanty, ter Hoeve,
  Burtsev, Skrynnik, Zholus, Panov, Srinet et~al.}]{kiseleva2021iglu}
Julia Kiseleva, Ziming Li, Mohammad Aliannejadi, Shrestha Mohanty, Maartje ter
  Hoeve, Mikhail Burtsev, Alexey Skrynnik, Artem Zholus, Aleksandr Panov, Kavya
  Srinet, et~al. 2021.
\newblock Neurips 2021 competition iglu: Interactive grounded language
  understanding in a collaborative environment.
\newblock \emph{arXiv preprint arXiv:2110.06536}.

\bibitem[{Kiseleva et~al.(2022)Kiseleva, Skrynnik, Zholus, Mohanty, Arabzadeh,
  C{\^o}t{\'e}, Aliannejadi, Teruel, Li, Burtsev et~al.}]{kiseleva2022iglu}
Julia Kiseleva, Alexey Skrynnik, Artem Zholus, Shrestha Mohanty, Negar
  Arabzadeh, Marc-Alexandre C{\^o}t{\'e}, Mohammad Aliannejadi, Milagro Teruel,
  Ziming Li, Mikhail Burtsev, et~al. 2022.
\newblock Iglu 2022: Interactive grounded language understanding in a
  collaborative environment at neurips 2022.
\newblock \emph{arXiv preprint arXiv:2205.13771}.

\bibitem[{Koulouri and Lauria(2009)}]{koulouri-lauria-2009-exploring}
Theodora Koulouri and Stanislao Lauria. 2009.
\newblock \href {https://aclanthology.org/W09-3915} {Exploring miscommunication
  and collaborative behaviour in human-robot interaction}.
\newblock In \emph{Proceedings of the {SIGDIAL} 2009 Conference}, pages
  111--119, London, UK. Association for Computational Linguistics.

\bibitem[{Kumar and Black(2020)}]{kumar-black-2020-clarq}
Vaibhav Kumar and Alan~W Black. 2020.
\newblock \href {https://doi.org/10.18653/v1/2020.acl-main.651} {{C}lar{Q}: A
  large-scale and diverse dataset for clarification question generation}.
\newblock In \emph{Proceedings of the 58th Annual Meeting of the Association
  for Computational Linguistics}, pages 7296--7301, Online. Association for
  Computational Linguistics.

\bibitem[{Lambert et~al.(2019)Lambert, Cordes, Kaplan, Jayannavar, and
  Hockenmaier}]{lambert2019virtual}
Charlotte Lambert, Ariel Cordes, Elli Kaplan, Prashant Jayannavar, and Julia
  Hockenmaier. 2019.
\newblock \href {http://dreuarchive.cra.org/2019/Lambert/#final_report}
  {Virtual world context encoding for grounded dialogue in minecraft}.

\bibitem[{Lee et~al.(2021)Lee, Kim, Hur, and Lim}]{Lee2021VisualTO}
Hyunhee Lee, Gyeongmin Kim, Yuna Hur, and Heuiseok Lim. 2021.
\newblock Visual thinking of neural networks: Interactive text to image
  synthesis.
\newblock \emph{IEEE Access}, 9:64510--64523.

\bibitem[{Liu et~al.(2020)Liu, Deng, Li, Cai, Xu, Wang, and
  Huang}]{Liu2020IRGANIM}
Zhenhuan Liu, Jincan Deng, Liang Li, Shaofei Cai, Qianqian Xu, Shuhui Wang, and
  Qingming Huang. 2020.
\newblock Ir-gan: Image manipulation with linguistic instruction by increment
  reasoning.
\newblock \emph{Proceedings of the 28th ACM International Conference on
  Multimedia}.

\bibitem[{Majumder et~al.(2021)Majumder, Rao, Galley, and
  McAuley}]{majumder-etal-2021-ask}
Bodhisattwa~Prasad Majumder, Sudha Rao, Michel Galley, and Julian McAuley.
  2021.
\newblock \href {https://doi.org/10.18653/v1/2021.naacl-main.340} {Ask what{'}s
  missing and what{'}s useful: Improving clarification question generation
  using global knowledge}.
\newblock In \emph{Proceedings of the 2021 Conference of the North American
  Chapter of the Association for Computational Linguistics: Human Language
  Technologies}, pages 4300--4312, Online. Association for Computational
  Linguistics.

\bibitem[{Marge and Rudnicky(2015)}]{marge-rudnicky-2015-miscommunication}
Matthew Marge and Alexander Rudnicky. 2015.
\newblock \href {https://doi.org/10.18653/v1/W15-4604} {Miscommunication
  recovery in physically situated dialogue}.
\newblock In \emph{Proceedings of the 16th Annual Meeting of the Special
  Interest Group on Discourse and Dialogue}, pages 22--31, Prague, Czech
  Republic. Association for Computational Linguistics.

\bibitem[{Matsumori et~al.(2021)Matsumori, Abe, Shingyouchi, Sugiura, and
  Imai}]{Matsumori2021LatteGANVG}
Shoya Matsumori, Yukikoko Abe, Kosuke Shingyouchi, Komei Sugiura, and Michita
  Imai. 2021.
\newblock Lattegan: Visually guided language attention for multi-turn
  text-conditioned image manipulation.
\newblock \emph{IEEE Access}, 9:160521--160532.

\bibitem[{Mitsuda et~al.(2022)Mitsuda, Higashinaka, Oga, and
  Yoshida}]{mitsuda-etal-2022-dialogue}
Koh Mitsuda, Ryuichiro Higashinaka, Yuhei Oga, and Sen Yoshida. 2022.
\newblock \href {https://aclanthology.org/2022.lrec-1.618} {Dialogue collection
  for recording the process of building common ground in a collaborative task}.
\newblock In \emph{Proceedings of the Thirteenth Language Resources and
  Evaluation Conference}, pages 5749--5758, Marseille, France. European
  Language Resources Association.

\bibitem[{Mohanty et~al.(2022)Mohanty, Arabzadeh, Teruel, Sun, Zholus,
  Skrynnik, Burtsev, Srinet, Panov, Szlam, Côté, and Kiseleva}]{iglu2022data}
Shrestha Mohanty, Negar Arabzadeh, Milagro Teruel, Yuxuan Sun, Artem Zholus,
  Alexey Skrynnik, Mikhail Burtsev, Kavya Srinet, Aleksandr Panov, Arthur
  Szlam, Marc-Alexandre Côté, and Julia Kiseleva. 2022.
\newblock \href {https://doi.org/10.48550/ARXIV.2211.06552} {Collecting
  interactive multi-modal datasets for grounded language understanding}.

\bibitem[{Narayan-Chen et~al.(2019)Narayan-Chen, Jayannavar, and
  Hockenmaier}]{narayan-chen-etal-2019-collaborative}
Anjali Narayan-Chen, Prashant Jayannavar, and Julia Hockenmaier. 2019.
\newblock \href {https://doi.org/10.18653/v1/P19-1537} {Collaborative dialogue
  in {M}inecraft}.
\newblock In \emph{Proceedings of the 57th Annual Meeting of the Association
  for Computational Linguistics}, pages 5405--5415, Florence, Italy.
  Association for Computational Linguistics.

\bibitem[{Padmakumar et~al.(2022)Padmakumar, Thomason, Shrivastava, Lange,
  Narayan-Chen, Gella, Piramuthu, Tur, and Hakkani-Tur}]{padmakumar2022teach}
Aishwarya Padmakumar, Jesse Thomason, Ayush Shrivastava, Patrick Lange, Anjali
  Narayan-Chen, Spandana Gella, Robinson Piramuthu, Gokhan Tur, and Dilek
  Hakkani-Tur. 2022.
\newblock Teach: Task-driven embodied agents that chat.
\newblock In \emph{Proceedings of the AAAI Conference on Artificial
  Intelligence}, volume~36, pages 2017--2025.

\bibitem[{Purver et~al.(2003)Purver, Ginzburg, and Healey}]{purver2003means}
Matthew Purver, Jonathan Ginzburg, and Patrick Healey. 2003.
\newblock On the means for clarification in dialogue.
\newblock In \emph{Current and new directions in discourse and dialogue}, pages
  235--255. Springer.

\bibitem[{Rao and Daum{\'e}~III(2018)}]{rao-daume-iii-2018-learning}
Sudha Rao and Hal Daum{\'e}~III. 2018.
\newblock \href {https://doi.org/10.18653/v1/P18-1255} {Learning to ask good
  questions: Ranking clarification questions using neural expected value of
  perfect information}.
\newblock In \emph{Proceedings of the 56th Annual Meeting of the Association
  for Computational Linguistics (Volume 1: Long Papers)}, pages 2737--2746,
  Melbourne, Australia. Association for Computational Linguistics.

\bibitem[{Reimers and Gurevych(2019)}]{reimers-2019-sentence-bert}
Nils Reimers and Iryna Gurevych. 2019.
\newblock \href {http://arxiv.org/abs/1908.10084} {Sentence-bert: Sentence
  embeddings using siamese bert-networks}.
\newblock In \emph{Proceedings of the 2019 Conference on Empirical Methods in
  Natural Language Processing}. Association for Computational Linguistics.

\bibitem[{Rieser et~al.(2005)Rieser, Kruijff-Korbayov{\'a}, and
  Lemon}]{rieser-etal-2005-corpus}
Verena Rieser, Ivana Kruijff-Korbayov{\'a}, and Oliver Lemon. 2005.
\newblock \href {https://aclanthology.org/2005.sigdial-1.11} {A corpus
  collection and annotation framework for learning multimodal clarification
  strategies}.
\newblock In \emph{Proceedings of the 6th SIGdial Workshop on Discourse and
  Dialogue}, pages 97--106, Lisbon, Portugal. Special Interest Group on
  Discourse and Dialogue (SIGdial).

\bibitem[{Rieser and Lemon(2006)}]{rieser-lemon-2006-using}
Verena Rieser and Oliver Lemon. 2006.
\newblock \href {https://aclanthology.org/P06-2085} {Using machine learning to
  explore human multimodal clarification strategies}.
\newblock In \emph{Proceedings of the {COLING}/{ACL} 2006 Main Conference
  Poster Sessions}, pages 659--666, Sydney, Australia. Association for
  Computational Linguistics.

\bibitem[{Rieser and Moore(2005)}]{rieser-moore-2005-implications}
Verena Rieser and Johanna Moore. 2005.
\newblock \href {https://doi.org/10.3115/1219840.1219870} {Implications for
  generating clarification requests in task-oriented dialogues}.
\newblock In \emph{Proceedings of the 43rd Annual Meeting of the Association
  for Computational Linguistics ({ACL}{'}05)}, pages 239--246, Ann Arbor,
  Michigan. Association for Computational Linguistics.

\bibitem[{Rodr{\'\i}guez and Schlangen(2004)}]{rodriguez2004form}
Kepa~Joseba Rodr{\'\i}guez and David Schlangen. 2004.
\newblock Form, intonation and function of clarification requests in german
  task-oriented spoken dialogues.
\newblock In \emph{Proceedings of Catalog (the 8th workshop on the semantics
  and pragmatics of dialogue; SemDial04)}.

\bibitem[{Roth et~al.(2022)Roth, Anthonio, and Sauer}]{roth-etal-2022-semeval}
Michael Roth, Talita Anthonio, and Anna Sauer. 2022.
\newblock \href {https://doi.org/10.18653/v1/2022.semeval-1.146}
  {{S}em{E}val-2022 task 7: Identifying plausible clarifications of implicit
  and underspecified phrases in instructional texts}.
\newblock In \emph{Proceedings of the 16th International Workshop on Semantic
  Evaluation (SemEval-2022)}, pages 1039--1049, Seattle, United States.
  Association for Computational Linguistics.

\bibitem[{Saito and Rehmsmeier(2015)}]{saito2015precision}
Takaya Saito and Marc Rehmsmeier. 2015.
\newblock The precision-recall plot is more informative than the roc plot when
  evaluating binary classifiers on imbalanced datasets.
\newblock \emph{PloS one}, 10(3):e0118432.

\bibitem[{Schlangen(2004)}]{schlangen-2004-causes}
David Schlangen. 2004.
\newblock \href {https://aclanthology.org/W04-2325} {Causes and strategies for
  requesting clarification in dialogue}.
\newblock In \emph{Proceedings of the 5th {SIG}dial Workshop on Discourse and
  Dialogue at {HLT}-{NAACL} 2004}, pages 136--143, Cambridge, Massachusetts,
  USA. Association for Computational Linguistics.

\bibitem[{Schlangen and
  Fern{\'a}ndez(2007{\natexlab{a}})}]{schlangen-fernandez-2007-beyond}
David Schlangen and Raquel Fern{\'a}ndez. 2007{\natexlab{a}}.
\newblock \href {https://aclanthology.org/2007.sigdial-1.10} {Beyond repair
  {--} testing the limits of the conversational repair system}.
\newblock In \emph{Proceedings of the 8th SIGdial Workshop on Discourse and
  Dialogue}, pages 51--54, Antwerp, Belgium. Association for Computational
  Linguistics.

\bibitem[{Schlangen and
  Fern{\'a}ndez(2007{\natexlab{b}})}]{schlangen2007speaking}
David Schlangen and Raquel Fern{\'a}ndez. 2007{\natexlab{b}}.
\newblock Speaking through a noisy channel-experiments on inducing
  clarification behaviour in human-human dialogue.
\newblock \emph{Proceedings of Interspeech 2007}.

\bibitem[{Schl{\"o}der and
  Fern{\'a}ndez(2014)}]{schloder-fernandez-2014-clarification}
Julian Schl{\"o}der and Raquel Fern{\'a}ndez. 2014.
\newblock \href {http://semdial.org/anthology/Z14-Schlöder_semdial_0047.pdf}
  {Clarification requests at the level of uptake}.
\newblock In \emph{Proceedings of the 18th Workshop on the Semantics and
  Pragmatics of Dialogue - Poster Abstracts}, Edinburgh, United Kingdom.
  SEMDIAL.

\bibitem[{Schl{\"o}der and
  Fern{\'a}ndez(2015)}]{schloder-fernandez-2015-clarifying}
Julian~J. Schl{\"o}der and Raquel Fern{\'a}ndez. 2015.
\newblock \href {https://aclanthology.org/W15-0106} {Clarifying intentions in
  dialogue: A corpus study}.
\newblock In \emph{Proceedings of the 11th International Conference on
  Computational Semantics}, pages 46--51, London, UK. Association for
  Computational Linguistics.

\bibitem[{Shi et~al.(2022)Shi, Feng, and Lipani}]{shi-etal-2022-learning}
Zhengxiang Shi, Yue Feng, and Aldo Lipani. 2022.
\newblock \href {https://doi.org/10.18653/v1/2022.findings-naacl.158} {Learning
  to execute actions or ask clarification questions}.
\newblock In \emph{Findings of the Association for Computational Linguistics:
  NAACL 2022}, pages 2060--2070, Seattle, United States. Association for
  Computational Linguistics.

\bibitem[{Stoyanchev et~al.(2013)Stoyanchev, Liu, and
  Hirschberg}]{stoyanchev-etal-2013-modelling}
Svetlana Stoyanchev, Alex Liu, and Julia Hirschberg. 2013.
\newblock \href {https://aclanthology.org/W13-4021} {Modelling human
  clarification strategies}.
\newblock In \emph{Proceedings of the {SIGDIAL} 2013 Conference}, pages
  137--141, Metz, France. Association for Computational Linguistics.

\bibitem[{Stoyanchev et~al.(2014)Stoyanchev, Liu, and
  Hirschberg}]{stoyanchev2014towards}
Svetlana Stoyanchev, Alex Liu, and Julia Hirschberg. 2014.
\newblock Towards natural clarification questions in dialogue systems.
\newblock In \emph{AISB symposium on questions, discourse and dialogue},
  volume~20.

\bibitem[{Stoyanchev et~al.(2012)Stoyanchev, Liu, and
  Hirschberg}]{stoyanchev2012clarification}
Svetlana Stoyanchev, Alex Liu, and Julia~Bell Hirschberg. 2012.
\newblock Clarification questions with feedback.

\bibitem[{Thomason et~al.(2020)Thomason, Murray, Cakmak, and
  Zettlemoyer}]{thomason2020vision}
Jesse Thomason, Michael Murray, Maya Cakmak, and Luke Zettlemoyer. 2020.
\newblock Vision-and-dialog navigation.
\newblock In \emph{Conference on Robot Learning}, pages 394--406. PMLR.

\bibitem[{Zhang et~al.(2021)Zhang, Tseng, Jiang, Yang, Lee, and
  Essa}]{Zhang2021TextAN}
Tianhao Zhang, Hung-Yu Tseng, Lu~Jiang, Weilong Yang, Honglak Lee, and Irfan
  Essa. 2021.
\newblock Text as neural operator: Image manipulation by text instruction.
\newblock \emph{Proceedings of the 29th ACM International Conference on
  Multimedia}.

\bibitem[{Zitnick and Parikh(2013)}]{zitnick2013bringing}
C~Lawrence Zitnick and Devi Parikh. 2013.
\newblock Bringing semantics into focus using visual abstraction.
\newblock In \emph{Proceedings of the IEEE Conference on Computer Vision and
  Pattern Recognition}, pages 3009--3016.

\bibitem[{Zitnick et~al.(2013)Zitnick, Parikh, and
  Vanderwende}]{zitnick2013learning}
C~Lawrence Zitnick, Devi Parikh, and Lucy Vanderwende. 2013.
\newblock Learning the visual interpretation of sentences.
\newblock In \emph{Proceedings of the IEEE International Conference on Computer
  Vision}, pages 1681--1688.

\end{thebibliography}
